\documentclass{article}
\PassOptionsToPackage{numbers, compress}{natbib}
\usepackage[final,main]{neurips_2026}
\usepackage{graphicx}
\usepackage[utf8]{inputenc}
\usepackage[T1]{fontenc}
\usepackage{hyperref}
\usepackage{url}
\usepackage{booktabs}
\usepackage{amsfonts}
\usepackage{amsmath}
\usepackage{amssymb}
\usepackage{nicefrac}
\usepackage{microtype}
\usepackage{xcolor}
\usepackage{multirow}
\usepackage{longtable}
\usepackage{array}

\title{
  Modeling Whole-Slide Images as Dynamic Tumor Microenvironment Fields
}

\author{
\textbf{Lei Wu}\textsuperscript{1*},
\textbf{Jiashuai Liu}\textsuperscript{1*},
\textbf{Di Zhang}\textsuperscript{1},
\textbf{Zhangpeng Gong}\textsuperscript{1},
\textbf{Yingkang Zhan}\textsuperscript{1},
\textbf{Yi Niu}\textsuperscript{1},
\\
\textbf{Jiusong Ge}\textsuperscript{1},
\textbf{Chunze Yang}\textsuperscript{1},
\textbf{Kai Yi}\textsuperscript{3},
\textbf{Mireia Crispin-Ortuzar}\textsuperscript{3},
\textbf{Chen Li}\textsuperscript{2,1$\dagger$},
\textbf{Zeyu Gao}\textsuperscript{3$\dagger$}
\\[1.5mm]
{\normalfont\mdseries
\begin{tabular}{c}
\textsuperscript{1}School of Computer Science and Technology, Xi'an Jiaotong University\\
\textsuperscript{2}Institute of Intelligent Medicine, Chinese Academy of Medical Sciences\\
\textsuperscript{3}University of Cambridge
\end{tabular}
}
}

\begin{document}

\maketitle

\begingroup
\renewcommand{\thefootnote}{}
\footnotetext{%
\begin{tabular}{@{}ll@{}}
\textsuperscript{*} &
These authors contributed equally.\\
\textsuperscript{$\dagger$} &
Co-corresponding authors:
\texttt{cli@pumc.edu.cn, zg323@cam.ac.uk}
\end{tabular}
}
\endgroup

\begin{abstract}
Due to the gigapixel-scale nature of whole-slide images (WSIs), weakly supervised WSI analysis is commonly formulated as a multiple instance learning (MIL) problem, where patch-level features are aggregated into slide-level representations. However, diagnostic and prognostic evidence often arises from spatially coherent tumor microenvironment regions and their interactions, rather than isolated patches alone.
Existing patch-level or static region-based methods usually overlook how tissue regions should be adaptively formed and subsequently evolved through microenvironment interactions across heterogeneous boundaries.
In this paper, we propose \textbf{Concept-Guided Tumor Microenvironment Evolution} (\textbf{TMEvolve}), a reaction-diffusion-inspired framework that models WSIs as latent tumor microenvironment fields over discrete patch graphs. TMEvolve instantiates this view as a learnable graph-discretized evolution process over patch neighborhoods. It first forms adaptive soft tissue regions as coherent microenvironment units, then performs pseudo-time evolution through two complementary local dynamics: intra-region diffusion, which stabilizes latent states within coherent tissue compartments, and concept-guided boundary flux, which propagates visual feature signals and language-derived concept signals across heterogeneous region interfaces. The evolved microenvironment regions are finally aggregated for slide-level prediction.
We evaluate TMEvolve on six datasets across three weakly supervised WSI tasks: survival prediction, gene expression prediction, and histological subtype classification. TMEvolve consistently improves over representative MIL methods, pathology foundation models, and concept-guided baselines. Ablation studies and visualizations further support the effectiveness and interpretability of TMEvolve, highlighting the value of dynamic region modeling and boundary interaction.
\end{abstract}

\section{Introduction}

\begin{figure*}[!t]
\centering
\includegraphics[width=\textwidth]{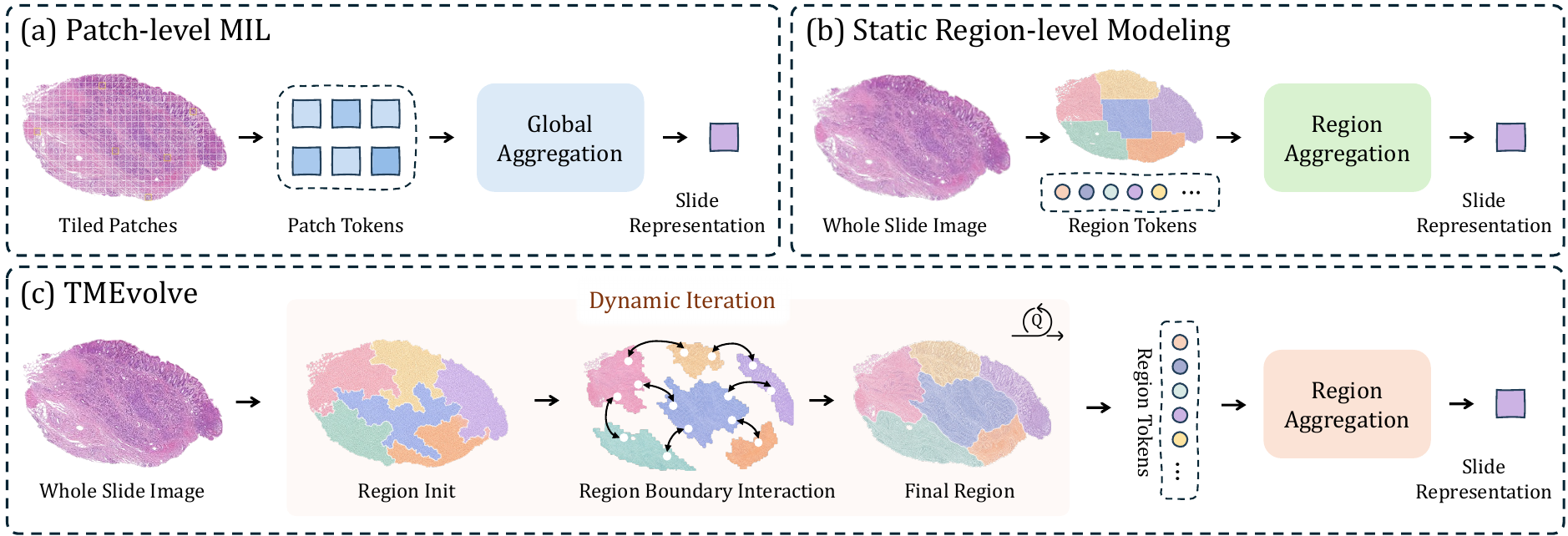}
\caption{
Overview of WSI modeling paradigms.
Patch-level MIL ignores region structure, while static region modeling relies on rigid regions.
TMEvolve models adaptive tissue regions and their concept-guided inter-region evolution.
}
\label{fig:intro}
\end{figure*}

Digitized whole-slide images provide gigapixel-scale histopathological morphology for computational pathology, supporting cancer diagnosis, prognosis, molecular prediction, and subtype classification~\cite{lu2021data,chen2020pathomic,chen2022pan,chen2024towards,zhang2026learnable}.
Since fine-grained annotations are expensive, weakly supervised multiple instance learning (MIL) is widely used, where each slide is represented as a bag of patches and trained with slide-level labels~\cite{pmlr-v80-ilse18a,lu2021data,li2021dual,shao2021transmil}. 
Recent pathology vision-language methods further enhance WSI representation learning by introducing semantic priors~\cite{huang2023visual,ikezogwo2023quilt,lu2024visual,zhao2024aligning,kapse2025gecko}. 
However, most existing WSI-level models still lack an explicit mechanism to organize patches into tissue-level structures and model interactions among microenvironment regions.

Conventional MIL directly aggregates patch tokens into a slide representation through attention pooling~\cite{pmlr-v80-ilse18a}, Transformer aggregation~\cite{shao2021transmil}, or other fusion mechanisms~\cite{li2021dual} (Figure~\ref{fig:intro}(a)).
Although effective, this patch-to-slide paradigm treats a WSI as a static collection of weakly related instances and overlooks region-level tissue organization. 
In routine diagnosis, pathologists often reason in a region-centric manner by organizing morphology into spatially coherent tissue compartments, such as tumor nests, stromal areas, and immune-enriched microenvironments. 
Static region modeling partially addresses this issue by aggregating local or pre-defined regions (Figure~\ref{fig:intro}(b))~\cite{chen2022scaling,chen2021whole,zheng2022graph,pmlr-v156-jaume21a,Cersovsky_2023_ICCV}. 
However, fixed grids, heuristic partitions, or static neighborhoods may cut across natural tissue boundaries and mix heterogeneous microenvironment components. 
Moreover, they usually treat regions as static aggregation units, without modeling how tissue states become coherent within regions or how heterogeneous regions influence each other across boundaries.

These limitations motivate a dynamic view of WSI representation learning. 
A WSI can be regarded as a spatial tissue field whose latent state varies over tissue coordinates. 
Within relatively homogeneous regions, neighboring patches tend to share coherent morphological and semantic states, suggesting a diffusion prior for local coordination~\cite{chen2021whole,pmlr-v156-jaume21a}.
Across heterogeneous interfaces, different microenvironment states may interact directionally, requiring concept-guided modeling of cross-region flux and reaction~\cite{de2023evolving,quail2013microenvironmental,friedl2011cancer,fridman2012immune,bindea2013spatiotemporal,joyce2015t}. 
This motivates a framework inspired by reaction--diffusion dynamics, where local states are stabilized within coherent tissue regions and semantic interactions are modeled across heterogeneous boundaries.

In this paper, we propose \textbf{Concept-Guided Tumor Microenvironment Evolution} (\textbf{TMEvolve}), a dynamic system for WSI representation learning informed by pathological concepts (Figure~\ref{fig:intro}(c)). 
Rather than solving a reaction--diffusion partial differential equation (PDE), TMEvolve instantiates this view as a learnable graph-discretized latent evolution framework over patch neighborhoods, motivated by studies linking graph neural message passing to diffusion or PDE-inspired dynamics~\cite{chamberlain2021grand,eliasof2021pde,chamberlain2021beltrami,wang2022acmp}. 
TMEvolve first forms adaptive soft tissue regions from morphological similarity and spatial continuity. 
It anchors these regions with language-encoded pathological concepts and evolves them in pseudo-time through intra-region diffusion and concept-guided boundary flux. 
The evolved region representations are finally aggregated for slide-level prediction.

We evaluate TMEvolve on survival prediction, gene expression prediction, and subtype classification. 
Across six cohorts, TMEvolve achieves strong performance compared with representative MIL methods, pathology foundation models, and survival-oriented baselines. 
Ablation studies validate the key design choices, while visualization results show that TMEvolve progressively refines tissue regions and provides interpretable evidence for slide-level prediction.

Our contributions are summarized as follows:

(1) We propose TMEvolve, an evolution-aware MIL framework that models WSIs as dynamic tumor microenvironment fields composed of interacting tissue regions. 
(2) We introduce dynamic region partition and intra-region coordination to construct spatially coherent and concept-bearing microenvironment units. 
(3) We design a concept-guided inter-region evolution mechanism to regulate boundary interactions between heterogeneous tissue regions. 
(4) TMEvolve achieves strong performance across survival prediction, gene expression prediction, and subtype classification, with ablation and visualization results supporting its effectiveness and interpretability.

Our code is included in the supplement, and the private cohort will be released upon acceptance.

\section{Related Work}

\noindent\textbf{Weakly supervised MIL baselines for WSI prediction.}
Weakly supervised multiple instance learning has become a standard paradigm for WSI analysis, where each slide is represented as a bag of patch instances and trained with slide-level labels. 
Representative MIL methods mainly differ in how patch-level information is aggregated into slide-level representations. 
ABMIL~\citep{pmlr-v80-ilse18a} introduces attention-based pooling for bag-level prediction, DSMIL~\citep{li2021dual} uses a dual-stream formulation to identify critical instances, TransMIL~\citep{shao2021transmil} models long-range patch correlations with Transformer aggregation, and RRT-MIL~\citep{tang2024feature} improves slide representations through feature re-embedding. 
These methods provide strong general baselines across survival prediction, gene expression prediction, and subtype classification. 
However, most of them primarily focus on patch aggregation, without explicitly modeling how heterogeneous tissue states interact during representation learning.

\noindent\textbf{Pathology foundation and vision-language models.}
Recent pathology foundation models provide stronger transferable representations for whole-slide analysis. 
Models such as CHIEF~\citep{wang2024pathology}, Feather~\citep{shao2025multiple}, and TITAN~\citep{ding2025multimodal} have shown strong performance in diagnosis, prognosis, and general WSI representation learning. 
Vision-language and concept-guided methods further introduce semantic priors into pathology modeling~\citep{gao2026alpaca,yang2026pathnavigate,ge2026thinking}. 
PLIP~\citep{huang2023visual}, Quilt~\citep{ikezogwo2023quilt}, and CONCH~\citep{lu2024visual} align pathology images with language representations, while CITE~\citep{zhang2023text}, QPMIL~\citep{gou2025queryable}, ConcepPath~\citep{zhao2024aligning}, and GECKO~\citep{kapse2025gecko} use textual concepts or vision-language cues for interpretable classification or molecular prediction. 
These methods are closely related to our comparisons because they use stronger pretrained representations or semantic guidance to improve downstream WSI prediction. 
Nevertheless, they mainly use language or concepts for global alignment, query-based aggregation, or concept-level prediction, whereas TMEvolve uses pathological concepts to guide interactions among tissue states during pseudo-time evolution.

\noindent\textbf{Spatial and microenvironment-aware WSI modeling.}
Beyond global slide-level aggregation, several studies have explored spatially aware WSI representations~\citep{gao2025smmile,zhang2026care}. 
Hierarchical, graph-based, or clustering-guided models, such as HIPT~\citep{chen2022scaling}, Patch-GCN~\citep{chen2021whole}, graph-based WSI modeling~\citep{zheng2022graph}, and clustering-guided graph MIL~\citep{pmlr-v156-jaume21a}, incorporate local tissue context or spatial relationships into slide representation learning. 
These methods show the value of spatial organization, but they usually use spatial structure as a fixed context for feature aggregation. 
From a biomedical perspective, tumor progression is shaped by the organization and interaction of tumor microenvironment components, including immune infiltration~\citep{galon2006type,fridman2012immune,bindea2013spatiotemporal}, tumor invasion~\citep{friedl2011cancer}, immune exclusion, and stromal remodeling~\citep{joyce2015t,kalluri2016biology}. 
Motivated by this view, TMEvolve models WSIs as dynamic tumor microenvironment fields, where concept-bearing tissue states are refined through intra-region coordination and concept-guided boundary interaction.

\section{Method}

\begin{figure*}[!t]
\centering
\includegraphics[width=\textwidth]{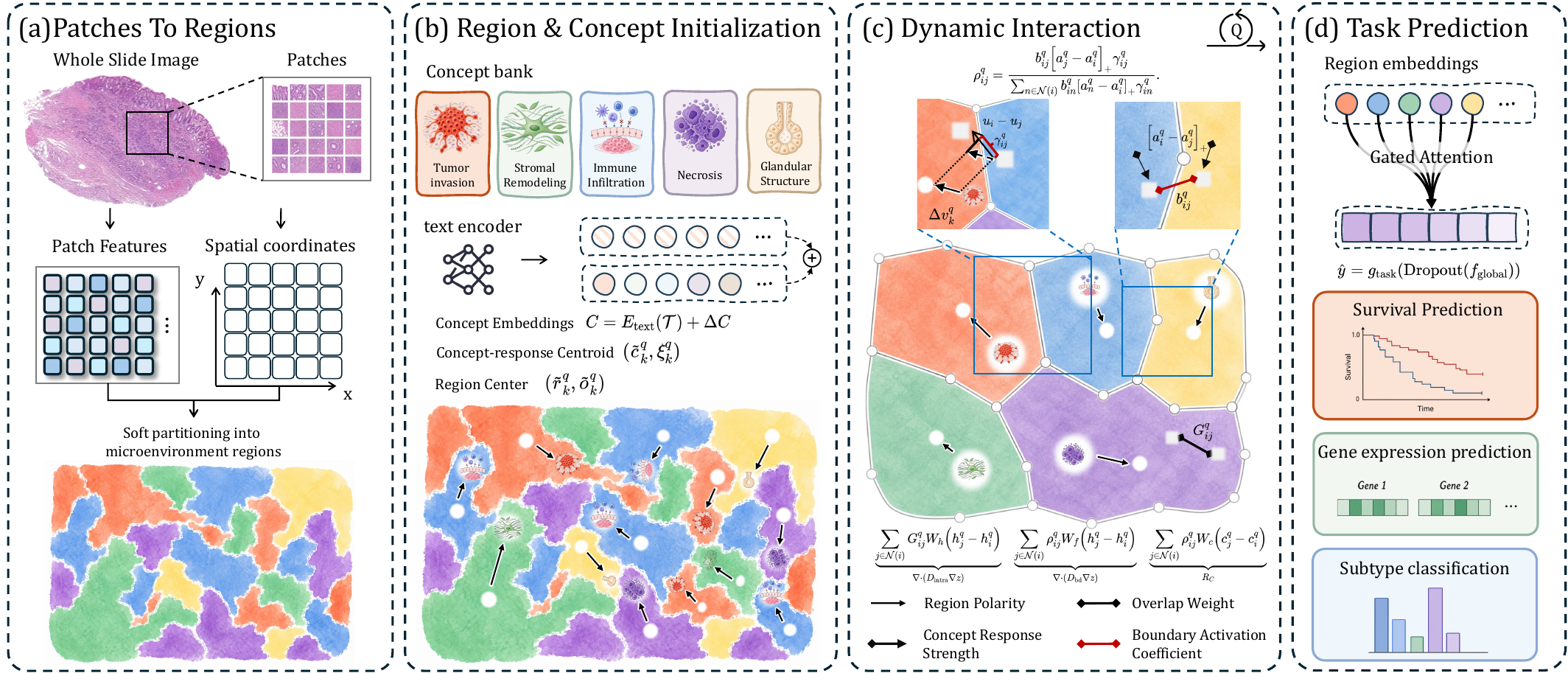}
\caption{
Overview of TMEvolve.
(a) TMEvolve first encodes WSI patches and spatial coordinates, and softly partitions them into adaptive regions.
(b) Region states are initialized with language-encoded concepts, concept-response centroids, and region centers.
(c) Dynamic interaction refines region representations through intra-region diffusion, concept-conditioned reaction, and anisotropic boundary flux.
(d) The evolved region embeddings are aggregated by gated attention for survival prediction, gene expression prediction, and subtype classification.
}
\label{fig:method_overview}
\end{figure*}

In this section, we present \textbf{TMEvolve}, an evolution-aware MIL framework that models a WSI as an adaptive tumor microenvironment field. 
Instead of aggregating static patch instances, TMEvolve formulates WSI representation learning as a graph-discretized latent reaction--diffusion process~\cite{wang2022acmp}, where within-region diffusion stabilizes coherent tissue compartments, boundary anisotropic flux models heterogeneous interface interactions, and concept-conditioned reaction incorporates language-encoded pathological semantics. 
As shown in Figure~\ref{fig:method_overview}, the following subsections describe the latent field formulation, adaptive region diffusion, concept-conditioned reaction, boundary flux, pseudo-time evolution, and task-specific aggregation.

\subsection{WSI as a Latent Reaction--Diffusion Microenvironment Field}

We first introduce the latent field formulation underlying TMEvolve. 
In a classical reaction--diffusion system, the evolution of a spatial state field $z(u,q)\in\mathbb{R}^d$ is written as~\cite{murray2002mathematical}
\begin{equation}
\frac{\partial z(u,q)}{\partial q}
=
\nabla\cdot\big(D(u,q)\nabla z(u,q)\big)
+
R(z,u,q),
\end{equation}
where $u$ denotes the spatial tissue coordinate, $q$ denotes the pseudo-time variable for iterative latent-state evolution, $D(u,q)$ controls spatial diffusion, and $R(z,u,q)$ denotes reaction effects.
For WSI representation learning, this formulation naturally matches the heterogeneous tissue field, where local consistency is needed within coherent regions and interactions are expected across heterogeneous interfaces.
We therefore decompose the diffusion operator into a within-region component $D_{\mathrm{intra}}$ and a boundary anisotropic component $D_{\mathrm{bd}}$, and introduce a concept-conditioned semantic source term $R_{\mathcal{C}}$:
\begin{equation}
\label{eq:latent_rd}
\frac{\partial z(u,q)}{\partial q}
=
\nabla\cdot\big(D_{\mathrm{intra}}(u,q)\nabla z(u,q)\big)
+
\nabla\cdot\big(D_{\mathrm{bd}}(u,q)\nabla z(u,q)\big)
+
R_{\mathcal{C}}(z,u,q).
\end{equation}
Here, the first term stabilizes representations inside coherent microenvironment regions, the second term models anisotropic flux across heterogeneous tissue interfaces, and the third term injects pathological concept semantics into latent state evolution. 
The pseudo-time variable $q$ denotes iterative representation refinement rather than chronological disease progression.

Since WSIs are observed as discrete patch sets, we implement this formulation on a patch neighborhood graph. 
Given initial patch features $H^0=[h_i^0]_{i=1}^{N}\in\mathbb{R}^{N\times d}$ and normalized coordinates $U=[u_i]_{i=1}^{N}\in\mathbb{R}^{N\times 2}$, the continuous state $z(u_i,q)$ is represented by $h_i^q$. 
A detailed graph-discretized derivation from Eq.~\eqref{eq:latent_rd} to the patch-level evolution updates is provided in Appendix~\ref{app:graph_discretization}.

\subsection{Adaptive Region Field and Within-region Diffusion}

We discretize the within-region diffusion term $\nabla\cdot(D_{\mathrm{intra}}\nabla z)$ by constructing an adaptive region field over WSI patches. 
The region field estimates where local information exchange should be encouraged: patches within coherent tissue compartments are allowed to diffuse, while propagation across heterogeneous or uncertain boundaries is suppressed.

At initialization, we obtain spatially dispersed region seeds by farthest point sampling. 
Each region $k$ is associated with a region feature prototype $r_k^q$ and a spatial center $o_k^q$. 
At pseudo-time step $q$, the adaptive region field is represented by a soft assignment matrix $S^q\in\mathbb{R}^{N\times K}$, where $S_{i,k}^q$ denotes the membership of patch $i$ to region $k$. 
To construct adaptive regions that are morphologically consistent and spatially continuous, the soft assignment is defined as:
\begin{equation}
E_{i,k}^q
=
\frac{1}{|\mathcal{N}(i)|}
\sum_{j\in\mathcal{N}(i)}
\left(
\left\|
\frac{h_j^q}{\|h_j^q\|_2}
-
\frac{r_k^q}{\|r_k^q\|_2}
\right\|_2^2
+
\|u_j-o_k^q\|_2^2
\right),
\quad
S_{i,k}^q
=
\frac{\exp(-E_{i,k}^q)}
{\sum_{k'=1}^{K}\exp(-E_{i,k'}^q)} .
\end{equation}
where $\mathcal{N}(i)$ denotes the spatial neighborhood of patch $i$.
The adaptive region field induces a state-dependent conductivity on the patch graph. 
For each neighboring patch pair $(i,j)$, we compute their normalized membership-overlap weight $G_{ij}^q$ and perform within-region diffusion as
\begin{equation}
G_{ij}^q
=
\frac{\sum_{k=1}^{K}S_{i,k}^qS_{j,k}^q}
{\sum_{n\in\mathcal{N}(i)}\sum_{k=1}^{K}S_{i,k}^qS_{n,k}^q},
\quad
\widetilde{h}_i^q
=
h_i^q
+
\sum_{j\in\mathcal{N}(i)}
G_{ij}^q W_h\left(h_j^q-h_i^q\right).
\end{equation}
Here, $G_{ij}^q$ serves as the state-dependent conductivity for $\nabla\cdot(D_{\mathrm{intra}}\nabla z)$, assigning larger weights to neighboring patches with similar region memberships and suppressing cross-boundary smoothing.

\subsection{Concept-guided Boundary Reaction--Flux Evolution}
After within-region diffusion, we further model interactions across heterogeneous microenvironment interfaces. 
Instead of treating boundary flux and concept reaction as two independent modules, TMEvolve couples them through a shared anisotropic boundary conductivity. 
This conductivity is derived from region heterogeneity, concept-response contrast, and region-level polarity.

For a given cancer type or downstream task, we collect task-relevant pathological descriptions $\mathcal{T}=\{\tau_1,\ldots,\tau_M\}$, such as invasive growth, immune infiltration, necrosis, and stromal remodeling. 
A text encoder maps these descriptions into concept embeddings, which are adapted by a learnable residual:
\begin{equation}
C = E_{\mathrm{text}}(\mathcal{T}) + \Delta C
= [c_1,\ldots,c_M]^\top \in \mathbb{R}^{M\times d}.
\end{equation}

Given the soft assignment matrix $S^q$ and the locally coordinated patch representations $\widetilde{H}^q$, 
we summarize each region by its semantic state 
$\widetilde{r}_k^q=\sum_{i=1}^{N}S_{i,k}^q\widetilde{h}_i^q/\sum_{i=1}^{N}S_{i,k}^q$ 
and geometric center 
$\widetilde{o}_k^q=\sum_{i=1}^{N}S_{i,k}^q u_i/\sum_{i=1}^{N}S_{i,k}^q$. 
Each region is then anchored to its dominant pathological concept by 
$\theta_k^q=\arg\max_m(\widetilde{r}_k^{q\top}c_m/\sqrt{d})$, 
and we set $\widetilde{c}_k^q=c_{\theta_k^q}$. 
The assigned concept $\widetilde{c}_k^q$ defines the region-level semantic identity and provides the semantic reference for locating concept-responsive patches within the region.

\noindent\textbf{Region polarity.}
To capture where the dominant concept is spatially concentrated inside region $k$, we compute the concept-response centroid and its deviation from the geometric region center as
\begin{equation}
\eta_{i,k}^q
=
\frac{
S_{i,k}^q
\exp\left(\widetilde{h}_i^{q\top}\widetilde{c}_k^q/\sqrt{d}\right)
}{
\sum_{\ell=1}^{N}
S_{\ell,k}^q
\exp\left(\widetilde{h}_\ell^{q\top}\widetilde{c}_k^q/\sqrt{d}\right)
},
\qquad
\xi_k^q=\sum_{i=1}^{N}\eta_{i,k}^q u_i,
\qquad
\Delta v_k^q=\xi_k^q-\widetilde{o}_k^q .
\end{equation}
Here, $\eta_{i,k}^q$ combines region membership and concept similarity, $\xi_k^q$ denotes the concept-response centroid, and $\Delta v_k^q$ captures its spatial bias relative to the geometric region center, providing the directional cue for anisotropic boundary flux.

\noindent\textbf{Concept response strength.}
We further assign the region-level concept back to patch-level states. 
For each patch $i$, we identify its dominant region by $\beta_i^q=\arg\max_k S_{i,k}^q$ and assign the corresponding concept $c_i^q=\widetilde{c}_{\beta_i^q}^q$. 
The patch-level concept response strength is defined as $a_i^q=\widetilde{h}_i^{q\top}c_i^q/\sqrt{d}$. 
The response strength $a_i^q$ is used to measure concept-response contrast between neighboring patches, which later regulates boundary interactions across heterogeneous region interfaces.

Many pathology-relevant signals arise around tumor--stroma, tumor--immune, 
glandular--stromal, and necrosis-related boundaries. 
Therefore, based on the region polarity $\Delta v_k^q$ and patch-level concept response strength $a_i^q$ defined above, 
we model cross-region interactions as anisotropic boundary reaction--flux. 
The boundary conductivity is jointly determined by \textbf{boundary heterogeneity}, \textbf{concept-response contrast}, and \textbf{directional compatibility}.

\noindent\textbf{Boundary heterogeneity.}
For each graph edge $(i,j)$, we define the boundary activation coefficient as 
$b_{ij}^q=1-\sum_{k=1}^{K}S_{i,k}^qS_{j,k}^q$. 
It measures the cross-region heterogeneity from soft assignments, where a larger $b_{ij}^q$ indicates that the edge is more likely to lie on a heterogeneous microenvironment boundary.

\noindent\textbf{Directional compatibility.}
To regulate the direction of boundary flux, we use the normalized polarity of the source patch region. 
For patch $j$, we identify its dominant region by 
$\epsilon_j^q=\arg\max_k S_{j,k}^q$ and define 
$\widehat{\Delta v}_j^q=\Delta v_{\epsilon_j^q}^q/\|\Delta v_{\epsilon_j^q}^q\|_2$. 
The directional compatibility coefficient is defined as 
$\gamma_{ij}^q=[(u_i-u_j)^\top\widehat{\Delta v}_j^q]_+$, where $[x]_+=\max(x,0)$. 
A larger $\gamma_{ij}^q$ indicates that the direction from source patch $j$ to target patch $i$ is more consistent with the concept-response polarity of the source region.

\noindent\textbf{Boundary conductivity.}
Together with the concept response strength $a_i^q$, we define the anisotropic boundary conductivity as
\begin{equation}
\rho_{ij}^q =
\frac{
b_{ij}^q\left[a_j^q-a_i^q\right]_+\gamma_{ij}^q
}{
\sum_{n\in\mathcal{N}(i)} b_{in}^q\left[a_n^q-a_i^q\right]_+\gamma_{in}^q
}.
\end{equation}
Here, $b_{ij}^q$ captures \textbf{boundary heterogeneity}, 
$\left[a_j^q-a_i^q\right]_+$ captures positive \textbf{concept-response contrast}, 
and $\gamma_{ij}^q$ imposes anisotropic \textbf{directional compatibility}. 
Thus, $\rho_{ij}^q$ serves as the graph-discretized conductivity for boundary-localized reaction--flux.

\noindent\textbf{Reaction--flux update.}
With $\rho_{ij}^q$, we instantiate the boundary anisotropic diffusion term 
$\nabla\cdot(D_{\mathrm{bd}}\nabla z)$ and the concept-conditioned semantic source term $R_{\mathcal{C}}$ on heterogeneous tissue interfaces. 
Since the locally coordinated representation $\widetilde{h}_i^q$ already contains the within-region diffusion update, the subsequent boundary-localized evolution is written as
\begin{equation} 
h_i^{q+1} 
= 
\widetilde{h}_i^q
+ 
\underbrace{ 
\sum_{j\in\mathcal{N}(i)}\rho_{ij}^q W_f\left(h_j^q-h_i^q\right) 
}_{\nabla\cdot(D_{\mathrm{bd}}\nabla z)} 
+ 
\underbrace{ 
\sum_{j\in\mathcal{N}(i)}\rho_{ij}^q W_c\left(c_j^q-c_i^q\right) 
}_{R_{\mathcal{C}}}. 
\end{equation}
The first boundary term models feature-level anisotropic flux across heterogeneous interfaces, 
and the second term instantiates $R_{\mathcal{C}}$ as a boundary-localized semantic reaction. 
Equivalently, by expanding $\widetilde{h}_i^q$, this update gives a graph-discretized realization of Eq.~\eqref{eq:latent_rd}, 
where the diffusion conductivities, concept responses, and region assignments are dynamically recomputed along pseudo-time.

\noindent\textbf{Region refresh.}
After each pseudo-time step, the updated patch states induce a new adaptive region field. 
Specifically, we recompute the patch-to-region soft assignment $S^{q+1}$ using the updated patch states and update the region prototype and spatial center as
\begin{equation}
r_k^{q+1}
=
\frac{\sum_{i=1}^{N}S_{i,k}^{q+1}h_i^{q+1}}{\sum_{i=1}^{N}S_{i,k}^{q+1}},
\quad
o_k^{q+1}
=
\frac{\sum_{i=1}^{N}S_{i,k}^{q+1}u_i}{\sum_{i=1}^{N}S_{i,k}^{q+1}}.
\end{equation}
Low-mass regions are also merged as described in the adaptive region construction step. 
Since $G_{ij}^q$, $\rho_{ij}^q$, $c_i^q$, and $S^q$ are recomputed from the current states, 
TMEvolve performs state-dependent latent field evolution rather than static graph message passing. 
After $Q$ evolution steps, we obtain the evolved patch states $H^Q$, the final soft assignment matrix $S^Q$, 
and the evolved region states $R^Q=\{r_k^Q\}_{k=1}^{K}$ from the final region update.

\subsection{Evolution-aware Aggregation and Task Prediction}

We aggregate the evolved region states for downstream prediction~\cite{pmlr-v80-ilse18a}. 
Each $r_k^Q$ encodes region-level morphology together with the accumulated effects of within-region diffusion, boundary anisotropic flux, and concept-conditioned reaction. 
The slide-level representation and prediction are computed by
\begin{equation}
f_{\mathrm{global}}
=
\sum_{k=1}^{K}\beta_k r_k^Q,
\quad
\hat{y}
=
g_{\mathrm{task}}\left(\mathrm{Dropout}(f_{\mathrm{global}})\right),
\end{equation}
where $\beta_k$ is the attention weight of region $k$, and $g_{\mathrm{task}}(\cdot)$ is the task-specific prediction head for survival prediction, subtype classification, or gene expression prediction.

To avoid degenerate region usage, we regularize the final region mass $\pi_k^Q=\frac{1}{N}\sum_{i=1}^{N}S_{i,k}^Q$ and define the overall objective as
\begin{equation}
\mathcal{L}_{\mathrm{div}}
=
\sum_{k=1}^{K}
\left(
\pi_k^Q-\frac{1}{K}
\right)^2,
\quad
\mathcal{L}
=
\mathcal{L}_{\mathrm{task}}
+
\lambda_{\mathrm{div}}\mathcal{L}_{\mathrm{div}}.
\end{equation}
Here, $\lambda_{\mathrm{div}}$ controls the strength of the diversity regularization, and $\mathcal{L}_{\mathrm{task}}$ is instantiated according to the downstream task.

\section{Experiments}

\subsection{Experimental Setup}

\noindent\textbf{Datasets and tasks.}
We evaluate TMEvolve on six datasets across survival prediction, gene expression prediction, and subtype classification. 
For public TCGA cohorts, we use LUAD~\cite{cancer2014comprehensive}, BLCA~\cite{cancer2014comprehensive}, and BRCA~\cite{cancer2012comprehensive}. 
Survival prediction is evaluated on these three TCGA cohorts and an additional private gallbladder cancer (GBC) cohort, which will be made publicly available in the future.
For gene expression prediction, we use LUAD, BLCA, and BRCA to predict eight clinically meaningful genes per cohort. 
For subtype classification, we evaluate on BRACS~\cite{brancati2022bracs} and EBRAINS~\cite{roetzer2022digital} under both fine-grained and coarse-grained settings. 

\noindent\textbf{Implementation details.}
We use CONCH v1.5~\cite{lu2024visual} as the patch feature extractor and Qwen3-Embedding-8B~\cite{zhang2025qwen3} as the text feature extractor. 
Pathological concepts are generated with GPT-5~\cite{openai2025gpt5} and reviewed by pathology experts. 
We set the diversity regularization weight to $\lambda_{\mathrm{div}}=0.1$. 
More dataset details and hyperparameter settings are provided in Appendix~\ref{app:experimental_details}, computational cost in Appendix~\ref{app:computational_cost}, and full concept banks in Appendix~\ref{app:concept_construction}.

\noindent\textbf{Evaluation metrics.}
We report the concordance index (C-index) for survival prediction, Pearson correlation coefficient for gene expression prediction, and AUC and accuracy (ACC) for subtype classification under both fine-grained and coarse-grained settings.

\begin{table*}[!t]
\centering
\caption{Main results on survival prediction. Results are C-index ($\times 100$), reported as mean $\pm$ std. The best result in each cohort is shown in bold and the second-best result is underlined.}
\label{tab:survival_results}
\vspace{1mm}
\scriptsize
\setlength{\tabcolsep}{2.5pt}
\renewcommand{\arraystretch}{1.05}
\resizebox{\textwidth}{!}{
\begin{tabular}{lcccccccccc}
\toprule
Cohort 
& ABMIL 
& DSMIL 
& TransMIL 
& RRT-MIL 
& CHIEF 
& Feather 
& TITAN 
& VLSA 
& ProtoSurv 
& Ours \\
\midrule
LUAD 
& 63.9$\pm$5.3 
& 64.5$\pm$3.3 
& \underline{66.8$\pm$4.8}
& 66.1$\pm$5.2 
& 63.1$\pm$3.0 
& 63.5$\pm$1.4 
& \underline{66.8$\pm$4.0}
& 64.5$\pm$4.8 
& 66.5$\pm$5.7 
& \textbf{69.5$\pm$4.2} \\

GBC 
& 75.0$\pm$4.8 
& 73.8$\pm$5.0 
& 73.3$\pm$3.6 
& 74.2$\pm$4.3 
& 73.8$\pm$5.3 
& 75.2$\pm$2.6 
& 75.4$\pm$3.5 
& 76.1$\pm$4.4 
& \underline{76.2$\pm$1.8}
& \textbf{77.4$\pm$3.8} \\

BLCA 
& 61.9$\pm$6.4 
& 55.5$\pm$7.1 
& 60.2$\pm$3.9 
& 61.6$\pm$4.1 
& 61.2$\pm$6.5 
& 58.6$\pm$3.8 
& \underline{63.0$\pm$2.0}
& 60.9$\pm$3.6 
& 62.6$\pm$7.0 
& \textbf{65.9$\pm$5.7} \\

BRCA 
& 66.7$\pm$4.2 
& 60.6$\pm$2.7 
& 66.1$\pm$3.8 
& 66.0$\pm$3.6 
& 66.8$\pm$3.1 
& 62.7$\pm$1.1 
& 67.4$\pm$3.7 
& 69.3$\pm$6.4 
& \textbf{70.4$\pm$1.1} 
& \underline{70.3$\pm$3.1} \\
\bottomrule
\end{tabular}
}
\vspace{-2mm}
\end{table*}

\begin{table*}[!t]
\centering
\caption{Main results on fine-grained and coarse-grained subtype classification. Results are reported as mean $\pm$ std. AUC and ACC are reported in percentage (\%). The best result in each metric is shown in bold and the second-best result is underlined.}
\label{tab:subtype_results}
\vspace{1mm}
\scriptsize
\setlength{\tabcolsep}{1.6pt}
\renewcommand{\arraystretch}{1.06}
\resizebox{\textwidth}{!}{
\begin{tabular}{lllcccccccccc}
\toprule
Dataset 
& Level
& Metric
& ABMIL 
& DSMIL 
& TransMIL 
& RRT-MIL 
& CHIEF 
& Feather 
& TITAN 
& CITE 
& QPMIL 
& Ours \\
\midrule
\multirow{4}{*}{EBRAINS}
& \multirow{2}{*}{Fine}
& AUC
& 97.9$\pm$0.2 
& 97.9$\pm$0.3 
& 98.0$\pm$0.3 
& 97.7$\pm$0.6 
& 97.2$\pm$0.4 
& 97.7$\pm$0.4 
& \underline{98.2$\pm$0.4}
& 98.0$\pm$0.5 
& 97.7$\pm$0.2 
& \textbf{98.8$\pm$0.2} \\

& 
& ACC
& 75.9$\pm$1.1 
& 76.8$\pm$1.5 
& 76.5$\pm$1.2 
& 76.3$\pm$2.2 
& 71.5$\pm$1.5 
& 76.4$\pm$2.3 
& \textbf{79.2$\pm$1.8} 
& 76.2$\pm$2.9 
& 75.7$\pm$2.2 
& \underline{79.0$\pm$1.3} \\

\cmidrule(lr){2-13}
& \multirow{2}{*}{Coarse}
& AUC
& 99.4$\pm$0.2
& 99.5$\pm$0.1
& 99.5$\pm$0.1
& 99.5$\pm$0.2
& 98.5$\pm$0.3
& 98.9$\pm$0.2
& 99.5$\pm$0.1
& 98.8$\pm$0.4
& \underline{99.6$\pm$0.1}
& \textbf{99.6$\pm$0.2} \\

&
& ACC
& 92.9$\pm$1.2
& 93.1$\pm$0.5
& 93.1$\pm$0.8
& 92.8$\pm$1.4
& 87.4$\pm$1.2
& 91.7$\pm$1.0
& \textbf{93.8$\pm$1.0}
& 92.0$\pm$0.8
& 93.5$\pm$0.4
& \underline{93.7$\pm$0.7} \\

\midrule
\multirow{4}{*}{BRACS}
& \multirow{2}{*}{Fine}
& AUC
& 85.7$\pm$0.7 
& 85.3$\pm$2.0 
& 85.4$\pm$1.3 
& 85.9$\pm$0.9 
& 86.3$\pm$1.5 
& \underline{88.0$\pm$2.3}
& 85.5$\pm$3.2 
& 86.6$\pm$2.3 
& 87.3$\pm$1.6 
& \textbf{89.4$\pm$0.6} \\

& 
& ACC
& 60.7$\pm$2.0 
& 59.4$\pm$3.9 
& 58.6$\pm$1.7 
& 58.6$\pm$1.6 
& \underline{62.7$\pm$3.9}
& 62.2$\pm$3.3 
& 61.4$\pm$5.0 
& 61.8$\pm$3.9 
& 61.5$\pm$2.0 
& \textbf{65.4$\pm$1.0} \\

\cmidrule(lr){2-13}
& \multirow{2}{*}{Coarse}
& AUC
& 92.0$\pm$2.3
& 90.2$\pm$3.5
& 90.8$\pm$3.6
& 90.9$\pm$2.6
& \underline{93.1$\pm$1.3}
& 91.9$\pm$1.3
& \textbf{93.5$\pm$1.7}
& 89.8$\pm$2.2
& 92.3$\pm$1.3
& 92.6$\pm$0.7 \\

&
& ACC
& 82.5$\pm$1.4
& 81.8$\pm$3.6
& 80.7$\pm$2.8
& 79.0$\pm$4.5
& 81.8$\pm$2.1
& 81.4$\pm$2.2
& 82.5$\pm$2.4
& 81.6$\pm$1.8
& \underline{83.3$\pm$2.8}
& \textbf{83.5$\pm$1.8} \\

\bottomrule
\end{tabular}
}
\vspace{-2mm}
\end{table*}

\begin{figure*}[!t]
\centering
\includegraphics[width=\textwidth]{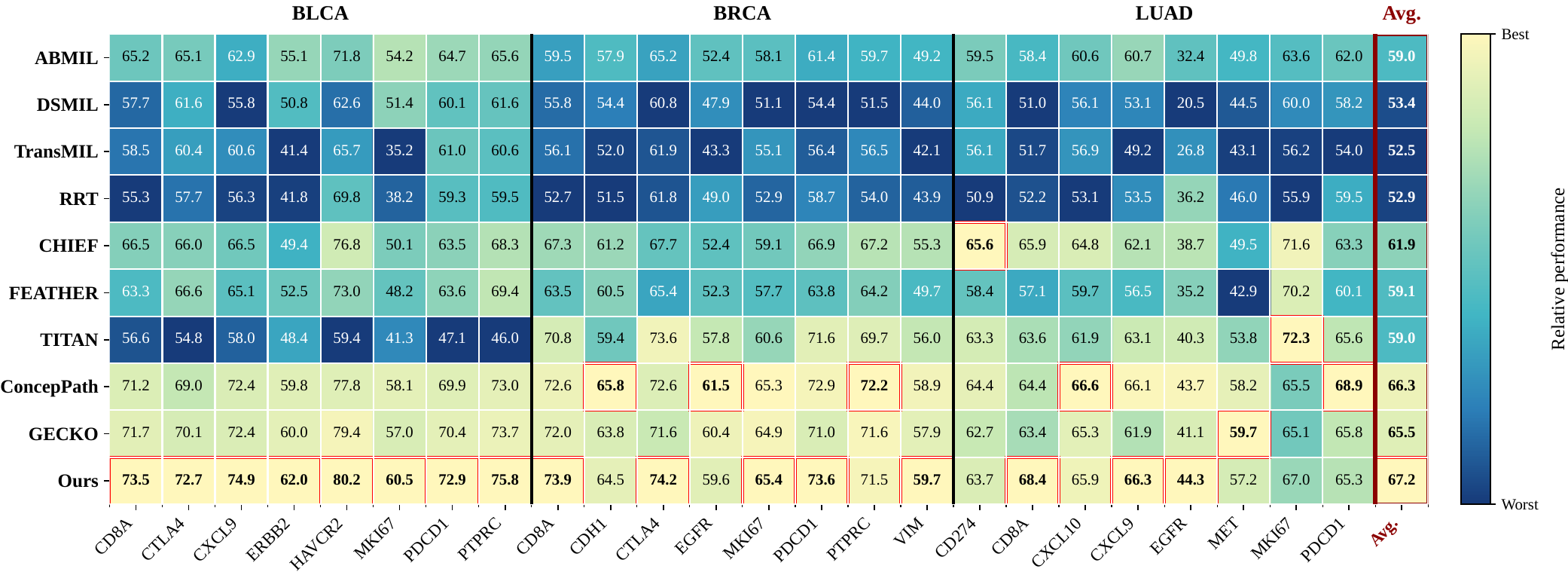}
\caption{
Gene-level prediction performance across TCGA datasets using Pearson correlation multiplied by 100.
Columns denote gene prediction tasks, rows denote compared methods, and color intensity indicates relative performance within each task, with brighter colors representing better results.
Red boxes highlight the best result in each column.
The average column summarizes overall performance across all gene-level tasks.
}
\label{fig:gene_results}
\vspace{-2mm}
\end{figure*}
\subsection{Comparison with State-of-the-art Methods}

\noindent\textbf{Comparison methods.}
We compare TMEvolve with representative WSI-level methods across survival prediction, subtype classification, and gene expression prediction. 
The compared methods include general MIL baselines such as ABMIL~\cite{pmlr-v80-ilse18a}, DSMIL~\cite{li2021dual}, TransMIL~\cite{shao2021transmil}, and RRT-MIL~\cite{tang2024feature}; pathology foundation or WSI-level models such as CHIEF~\cite{wang2024pathology}, Feather~\cite{shao2025multiple}, and TITAN~\cite{ding2025multimodal}; survival-oriented methods such as VLSA~\cite{liu2024interpretable} and ProtoSurv~\cite{wu2024leveraging}; subtype-oriented concept models such as CITE~\cite{zhang2023text} and QPMIL~\cite{gou2025queryable}; and gene-oriented concept-guided methods such as ConcepPath~\cite{zhao2024aligning} and GECKO~\cite{kapse2025gecko} when applicable.

\noindent\textbf{Main results.}
For \textit{survival prediction}, Table~\ref{tab:survival_results} shows that TMEvolve achieves the best performance on three out of four cohorts and remains highly competitive on BRCA. 
Compared with representative MIL, foundation-model, and survival-oriented baselines, TMEvolve shows consistent advantages on LUAD, GBC, and BLCA, suggesting that heterogeneous tissue regions and their interactions provide informative prognostic evidence. 
For \textit{subtype classification}, Table~\ref{tab:subtype_results} shows that TMEvolve performs strongly under both fine-grained and coarse-grained settings, achieving the best or competitive results on EBRAINS and BRACS across AUC and ACC. 
For \textit{gene expression prediction}, Figure~\ref{fig:gene_results} presents gene-level Pearson correlation results across TCGA cohorts. 
TMEvolve achieves strong and consistent performance on clinically relevant genes, especially immune- and proliferation-related genes, suggesting that dynamic microenvironment modeling helps capture weak and spatially distributed morphology-related molecular signals. 
Additional detailed gene-level results are provided in Appendix~\ref{app:gene_results}.

\subsection{Ablation Study and Hyperparameter Sensitivity}

\noindent\textbf{Main module ablation.}
As shown in Table~\ref{tab:ablation_results}, removing \textit{intra-region diffusion}, \textit{pseudo-time evolution}, or the diversity loss $\mathcal{L}_{\mathrm{div}}$ consistently degrades performance, demonstrating the importance of local diffusion, iterative refinement, and balanced region usage for stable microenvironment modeling. 
Replacing \textit{language-encoded concepts} with random embeddings also weakens the results, especially for gene expression prediction, suggesting that language-derived semantic priors help capture molecularly related morphology. 
Overall, the full TMEvolve performs best across tasks, supporting the joint effectiveness of region stabilization, concept guidance, and pseudo-time evolution.
\begin{table*}[!t]
\centering
\caption{Ablation study of TMEvolve on survival prediction, gene expression prediction, and BRACS subtype classification. Results are reported as mean $\pm$ std. The best result in each column is shown in bold.}
\label{tab:ablation_results}
\vspace{1mm}
\scriptsize
\setlength{\tabcolsep}{2.5pt}
\renewcommand{\arraystretch}{1.08}
\resizebox{\textwidth}{!}{
\begin{tabular}{lcccccc}
\toprule
Variant
& \multicolumn{2}{c}{Survival}
& \multicolumn{2}{c}{Gene Prediction}
& \multicolumn{2}{c}{Subtype Classification} \\
\cmidrule(lr){2-3}\cmidrule(lr){4-5}\cmidrule(lr){6-7}
& BLCA C-index
& LUAD C-index
& BLCA Pearson
& LUAD Pearson
& BRACS AUC
& BRACS ACC \\
\midrule
w/o Intra-region Diffusion
& 0.6524$\pm$0.0549 
& 0.6812$\pm$0.0474 
& 0.7058$\pm$0.0323 
& 0.6088$\pm$0.0284 
& 0.8883$\pm$0.0142 
& 0.6431$\pm$0.0218 \\

Random Concepts
& 0.6477$\pm$0.0570 
& 0.6728$\pm$0.0621 
& 0.7016$\pm$0.0297 
& 0.6044$\pm$0.0307 
& 0.8873$\pm$0.0119 
& 0.6344$\pm$0.0095 \\

w/o Pseudo-time Evolution
& 0.6421$\pm$0.0468 
& 0.6746$\pm$0.0540 
& 0.7022$\pm$0.0267 
& 0.6061$\pm$0.0298 
& 0.8885$\pm$0.0085 
& 0.6470$\pm$0.0243 \\

w/o $\mathcal{L}_{\mathrm{div}}$
& 0.6538$\pm$0.0460 
& 0.6817$\pm$0.0471 
& 0.7104$\pm$0.0263 
& 0.6125$\pm$0.0361 
& 0.8894$\pm$0.0084 
& 0.6513$\pm$0.0196 \\

Full TMEvolve 
& \textbf{0.6589$\pm$0.0570} 
& \textbf{0.6952$\pm$0.0418} 
& \textbf{0.7158$\pm$0.0326} 
& \textbf{0.6226$\pm$0.0288} 
& \textbf{0.8940$\pm$0.0057} 
& \textbf{0.6544$\pm$0.0095} \\
\bottomrule
\end{tabular}
}
\vspace{-2mm}
\end{table*}

\noindent\textbf{Sensitivity analysis.}
As shown in Table~\ref{tab:hyperparameter_sensitivity}, subtype classification works well with fewer pseudo-time steps and coarser regions, whereas survival and gene expression prediction benefit from more refined microenvironment modeling due to weaker and more spatially distributed signals. 
Moderate settings, such as $Q=3$ and $K=6$ or $K=8$, generally perform better, while further increasing $Q$ or $K$ brings no consistent gain and may introduce redundant interactions or over-smoothing.

\begin{table*}[!t]
\centering
\caption{Hyperparameter sensitivity analysis of TMEvolve with respect to the number of pseudo-time evolution steps $Q$ and the number of adaptive regions $K$. Results are reported as mean $\pm$ std. The best result in each column is shown in bold.}
\label{tab:hyperparameter_sensitivity}
\vspace{1mm}
\scriptsize
\setlength{\tabcolsep}{2.5pt}
\renewcommand{\arraystretch}{1.08}
\resizebox{\textwidth}{!}{
\begin{tabular}{lcccccc}
\toprule
Setting
& \multicolumn{2}{c}{Survival}
& \multicolumn{2}{c}{Gene Prediction}
& \multicolumn{2}{c}{Subtype Classification} \\
\cmidrule(lr){2-3}\cmidrule(lr){4-5}\cmidrule(lr){6-7}
& BLCA C-index
& LUAD C-index
& BLCA Pearson
& LUAD Pearson
& BRACS AUC
& BRACS ACC \\
\midrule
$Q=1$ 
& 0.6491$\pm$0.0562 
& 0.6791$\pm$0.0483 
& 0.7123$\pm$0.0273 
& 0.6076$\pm$0.0330 
& 0.8876$\pm$0.0066 
& 0.6472$\pm$0.0312 \\

$Q=2$ 
& 0.6485$\pm$0.0328 
& 0.6803$\pm$0.0532 
& 0.7111$\pm$0.0270 
& 0.6174$\pm$0.0354 
& \textbf{0.8940$\pm$0.0057} 
& \textbf{0.6544$\pm$0.0095} \\

$Q=3$ 
& \textbf{0.6589$\pm$0.0570} 
& \textbf{0.6952$\pm$0.0418} 
& \textbf{0.7158$\pm$0.0326} 
& \textbf{0.6226$\pm$0.0288} 
& 0.8830$\pm$0.0124 
& 0.6386$\pm$0.0322 \\

$Q=4$ 
& 0.6523$\pm$0.0337 
& 0.6894$\pm$0.0419 
& 0.7130$\pm$0.0271 
& 0.6091$\pm$0.0388 
& 0.8785$\pm$0.0127 
& 0.6349$\pm$0.0341 \\
\midrule
$K=2$ 
& 0.6417$\pm$0.0553 
& 0.6614$\pm$0.0408 
& 0.7038$\pm$0.0302 
& 0.6118$\pm$0.0279 
& 0.8894$\pm$0.0123 
& 0.6433$\pm$0.0208 \\

$K=4$ 
& 0.6444$\pm$0.0362 
& 0.6878$\pm$0.0449 
& 0.7084$\pm$0.0277 
& 0.6135$\pm$0.0317 
& \textbf{0.8940$\pm$0.0057} 
& \textbf{0.6544$\pm$0.0095} \\

$K=6$ 
& 0.6429$\pm$0.0476 
& \textbf{0.6952$\pm$0.0418} 
& 0.7107$\pm$0.0280 
& \textbf{0.6226$\pm$0.0288} 
& 0.8892$\pm$0.0109 
& 0.6489$\pm$0.0284 \\

$K=8$ 
& \textbf{0.6589$\pm$0.0570} 
& 0.6850$\pm$0.0464 
& \textbf{0.7158$\pm$0.0326} 
& 0.6220$\pm$0.0260 
& 0.8876$\pm$0.0081 
& 0.6488$\pm$0.0363 \\

$K=10$ 
& 0.6414$\pm$0.0523 
& 0.6830$\pm$0.0603 
& 0.7087$\pm$0.0322 
& 0.6201$\pm$0.0239 
& 0.8867$\pm$0.0090 
& 0.6415$\pm$0.0199 \\
\bottomrule
\end{tabular}
}
\vspace{-2mm}
\end{table*}

\subsection{Interpretability Analysis}
\begin{figure*}[!t]
\centering
\includegraphics[width=\textwidth]{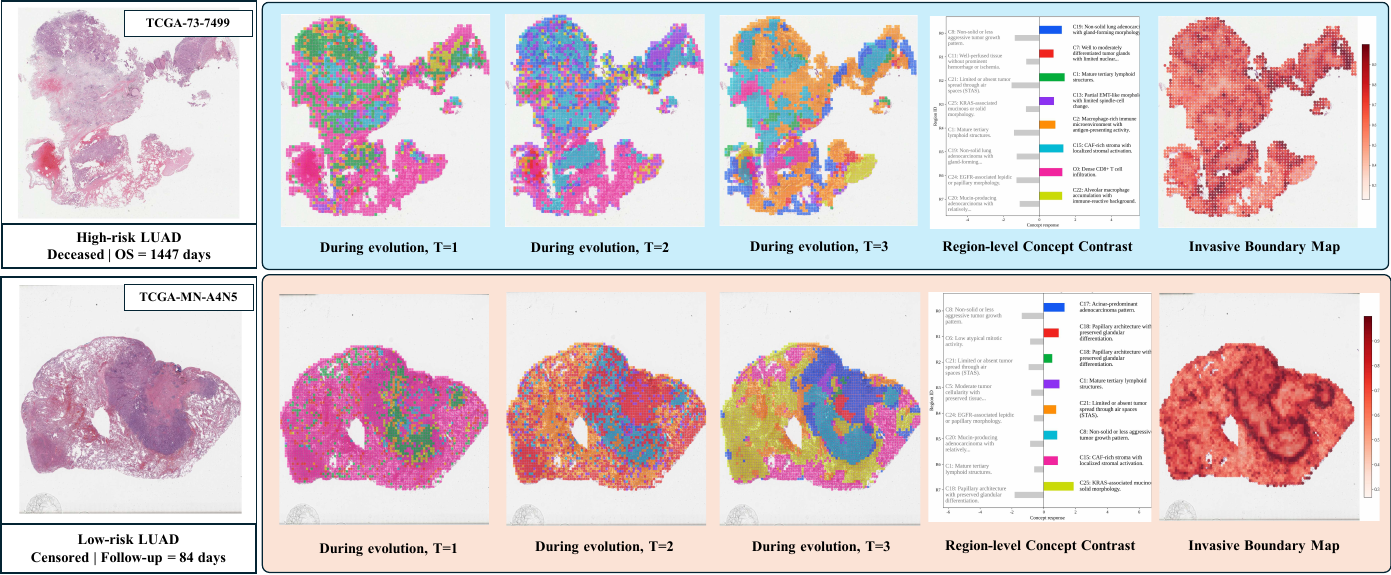}
\caption{
Microenvironment evolution in LUAD.
For each case, we show the original WSI thumbnail, region partitions during evolution, the region-level concept contrast, and the boundary response map.
The evolution process shows how TMEvolve progressively refines patch assignments into spatially coherent and concept-aligned microenvironment regions.
The concept contrast and boundary response map provide semantic and interaction-level evidence for slide-level risk prediction.
}
\label{fig:interpretability_vis}
\vspace{-2mm}
\end{figure*}

Beyond quantitative evaluation, we examine whether TMEvolve provides interpretable evidence for slide-level prediction. 
Figure~\ref{fig:interpretability_vis} visualizes representative LUAD cases with different risk outcomes. 
Rather than relying only on isolated high-attention patches, TMEvolve reveals how tissue organization is refined during pseudo-time evolution, where scattered assignments are gradually consolidated into spatially coherent and concept-aligned regions. 
The region-level concept contrast links these regions to clinically meaningful semantics, while the boundary response map localizes heterogeneous tissue interfaces involved in cross-region interaction. 
In high-risk cases, adverse tumor-related concept responses are more prominent, whereas low-risk cases show relatively stronger immune- or glandular-related patterns. 
Together, these visualizations provide complementary evidence for interpreting slide-level risk prediction, with additional cases provided in Appendix~\ref{app:visualization_results}.

\section{Conclusion}

In this paper, we present TMEvolve, an evolution-aware MIL framework that models WSIs as dynamic tumor microenvironment fields. 
By forming adaptive concept-bearing regions and evolving them across heterogeneous tissue boundaries, TMEvolve captures region-level microenvironment interactions beyond static patch aggregation. 
Experiments on survival prediction, gene expression prediction, and subtype classification demonstrate its effectiveness, while ablation and visualization analyses validate the contributions of intra-region diffusion, concept guidance, pseudo-time evolution, and interpretable boundary-level evidence. 
These results highlight dynamic microenvironment modeling as a promising direction for weakly supervised computational pathology.

\section*{Acknowledgments}

This work was supported by the Noncommunicable Chronic Diseases-National Science and Technology Major Project (2025ZD0544802),
the Shaanxi Province Key R\&D Program (2025SF-YBXM-363 and 2024SF-GJHX-32),
the Key Research and Development Program of Ningxia Hui Autonomous Region (2023BEG02023),
the ``Research on Key Technologies for Full-Chain Intelligent Pathological Diagnosis'' project of the First Affiliated Hospital of Xi'an Jiaotong University (HX202440),
the China Postdoctoral Science Foundation (2026M791685),
the National Natural Science Foundation of China (62506291),
the XJTU Research Fund for AI Science (2025YXYC004),
the Xi'an Jiaotong University Kunpeng \& Ascend Center of Cultivation,
the Fundamental Research Funds for the Central Universities (Nos.~xzy012026032 and xzy012026033),
the Royal Society (RGS\textbackslash R2\textbackslash252688),
and GE HealthCare.

\bibliographystyle{plainnat}
\bibliography{references}

\clearpage
\appendix

\begin{center}
\vspace*{4mm}

{\LARGE \bfseries Supplementary Material}

\vspace{2mm}

{\large Modeling Whole-Slide Images as Dynamic Tumor Microenvironment Fields}

\vspace{2mm}

{\normalsize Additional method details, extended experiments, and discussions for TMEvolve}

\vspace{5mm}
\end{center}

\vspace{3mm}

\noindent{\Large \bfseries Contents of the Supplement}

\vspace{2mm}

\begin{center}
\small
\setlength{\tabcolsep}{3pt}
\renewcommand{\arraystretch}{1.22}
\begin{tabular}{@{}p{0.10\textwidth}p{0.76\textwidth}r@{}}
\toprule
\textbf{Section} & \textbf{Content} & \textbf{Page} \\
\midrule

\textbf{A} & \textbf{Graph discretization of the latent reaction--diffusion formulation} 
& \pageref{app:graph_discretization} \\
\addlinespace[1.5mm]

\textbf{B} & \textbf{Additional experimental details} 
& \pageref{app:experimental_details} \\
B.1 & \hspace{0.8em}Dataset details 
& \pageref{app:dataset_details} \\
B.2 & \hspace{0.8em}Hyperparameter settings 
& \pageref{app:hyperparameter_settings} \\
\addlinespace[1.5mm]

\textbf{C} & \textbf{Computational cost analysis} 
& \pageref{app:computational_cost} \\
\addlinespace[1.5mm]

\textbf{D} & \textbf{Pathological concept construction} 
& \pageref{app:concept_construction} \\
\addlinespace[1.5mm]

\textbf{E} & \textbf{Additional gene expression prediction results} 
& \pageref{app:gene_results} \\
\addlinespace[1.5mm]

\textbf{F} & \textbf{Additional visualization results} 
& \pageref{app:visualization_results} \\
F.1 & \hspace{0.8em}Five-fold Kaplan--Meier survival curves 
& \pageref{app:km_curves} \\
F.2 & \hspace{0.8em}Additional interpretability visualization 
& \pageref{app:additional_interpretability} \\
\addlinespace[1.5mm]

\textbf{G} & \textbf{Limitations} 
& \pageref{app:limitations} \\
\addlinespace[1.5mm]

\textbf{H} & \textbf{Broader impact} 
& \pageref{app:broader_impact} \\

\bottomrule
\end{tabular}
\end{center}

\clearpage

\section{Graph Discretization of the Latent Reaction--Diffusion Formulation}
\label{app:graph_discretization}

We provide a graph-discretized derivation of the latent reaction--diffusion formulation used in TMEvolve. 
The goal is not to solve a continuous biological PDE, but to instantiate its structural form as a learnable latent evolution process over WSI patch neighborhoods.

Given patch locations $\{u_i\}_{i=1}^{N}$, we construct a spatial neighborhood graph with node set $\mathcal{V}$ and neighborhood $\mathcal{N}(i)$ for each patch $i$. 
The continuous latent state $z(u_i,q)$ at location $u_i$ is represented by the patch feature $h_i^q$. 
For a graph edge $(i,j)$, the spatial gradient of the latent field can be approximated by the edge-wise state difference
\begin{equation}
\nabla z(u_i,q)
\;\Longrightarrow\;
h_j^q-h_i^q.
\end{equation}
A state-dependent diffusion coefficient $D(u,q)$ is then represented by a graph conductivity $\kappa_{ij}^q$, which controls the strength of information exchange along edge $(i,j)$. 
Accordingly, the diffusion flux divergence at node $i$ is discretized as
\begin{equation}
\nabla\cdot(D\nabla z)(u_i,q)
\;\Longrightarrow\;
\sum_{j\in\mathcal{N}(i)}
\kappa_{ij}^q
\bigl(h_j^q-h_i^q\bigr).
\end{equation}

This discretization can also be interpreted as the negative gradient of a graph Dirichlet energy:
\begin{equation}
\mathcal{E}_{\kappa}(H^q)
=
\frac{1}{2}
\sum_{i=1}^{N}
\sum_{j\in\mathcal{N}(i)}
\kappa_{ij}^q
\left\|h_i^q-h_j^q\right\|_2^2.
\end{equation}
Taking a gradient descent step on this energy yields
\begin{equation}
-\frac{\partial \mathcal{E}_{\kappa}}{\partial h_i^q}
\propto
\sum_{j\in\mathcal{N}(i)}
\kappa_{ij}^q
\bigl(h_j^q-h_i^q\bigr),
\end{equation}
which corresponds to a graph diffusion update that encourages neighboring states with large conductivity to become more coherent.

In TMEvolve, the generic conductivity $\kappa_{ij}^q$ is instantiated in two complementary ways. 
For within-region diffusion, $\kappa_{ij}^q$ is implemented as the region-membership overlap weight $G_{ij}^q$, which assigns large conductivity to neighboring patches with similar adaptive region memberships. 
For boundary evolution, $\kappa_{ij}^q$ is implemented as the anisotropic boundary conductivity $\rho_{ij}^q$, which is activated by cross-region heterogeneity, concept-response contrast, and directional compatibility. 
Thus, the graph-discretized form of the latent reaction--diffusion equation becomes
\begin{equation}
h_i^{q+1}
=
h_i^q
+
\sum_{j\in\mathcal{N}(i)}
G_{ij}^q W_h
\bigl(h_j^q-h_i^q\bigr)
+
\sum_{j\in\mathcal{N}(i)}
\rho_{ij}^q W_f
\bigl(h_j^q-h_i^q\bigr)
+
\sum_{j\in\mathcal{N}(i)}
\rho_{ij}^q W_c
\bigl(c_j^q-c_i^q\bigr).
\end{equation}
The first summation discretizes within-region diffusion, the second summation discretizes boundary-localized anisotropic flux, and the third summation implements a concept-conditioned semantic reaction source. 
All conductivities and concept assignments are recomputed at each pseudo-time step, making the evolution process state-dependent rather than a fixed graph smoothing operation.
\section{Additional Experimental Details}
\label{app:experimental_details}

\subsection{Dataset Details}
\label{app:dataset_details}

Table~\ref{tab:dataset_details} summarizes the datasets used in our experiments. 
The evaluation covers three types of weakly supervised WSI tasks, including survival prediction, gene expression prediction, and subtype classification. 
For survival prediction, we use three TCGA cohorts, including LUAD, BLCA, and BRCA, together with a private GBC cohort. 
For gene expression prediction, we use TCGA LUAD, BLCA, and BRCA with matched RNA-seq profiles, where 25 genes are processed for each cohort~\cite{cancer2014comprehensive,cancer2012comprehensive}. 
For subtype classification, we evaluate TMEvolve on BRACS and EBRAINS under fine-grained and coarse-grained label settings~\cite{brancati2022bracs,roetzer2022digital}.

\begin{table*}[!h]
\centering
\caption{Summary of datasets used in our experiments.}
\label{tab:dataset_details}
\vspace{1mm}
\small
\setlength{\tabcolsep}{4pt}
\renewcommand{\arraystretch}{1.12}
\resizebox{\textwidth}{!}{
\begin{tabular}{lllll}
\toprule
Dataset & Source & Task & Label / Target & Dataset size / processed samples \\
\midrule
LUAD 
& TCGA 
& Survival prediction 
& Overall survival 
& 478 cases / 541 diagnostic slides \\

BLCA 
& TCGA 
& Survival prediction 
& Overall survival 
& 386 cases / 457 diagnostic slide files \\

BRCA 
& TCGA 
& Survival prediction 
& Overall survival 
& 1,098 cases / 1,133 FFPE slides \\

GBC 
& Private cohort 
& Survival prediction 
& Overall survival 
& 357 cases \\

\midrule
LUAD 
& TCGA 
& Gene expression prediction 
& 25 processed genes 
& 628 clinical samples; 623 matched with RNA-seq \\

BLCA 
& TCGA 
& Gene expression prediction 
& 25 processed genes 
& 435 clinical samples; 430 matched with RNA-seq \\

BRCA 
& TCGA 
& Gene expression prediction 
& 25 processed genes 
& 1,222 clinical samples; 1,220 matched with RNA-seq \\

\midrule
BRACS 
& Public dataset 
& Subtype classification 
& 7 fine / 3 coarse classes 
& 547 WSIs \\

EBRAINS 
& Digital Brain Tumour Atlas 
& Subtype classification 
& 30 fine / 12 coarse classes 
& 2,318 WSIs \\
\bottomrule
\end{tabular}
}
\vspace{-1mm}
\end{table*}

\subsection{Hyperparameter Settings}
\label{app:hyperparameter_settings}

Table~\ref{tab:main_hyperparams} summarizes the main implementation settings.
All experiments use five-fold cross-validation.
We use standard K-fold splitting for survival and gene expression prediction, and stratified K-fold splitting for subtype classification.
The dataset- and task-specific choices of the number of regions $K$ and evolution steps $Q$ are summarized in Table~\ref{tab:kt_settings}.

\begin{table}[!h]
\centering
\caption{Main hyperparameter settings of TMEvolve.}
\label{tab:main_hyperparams}
\vspace{1mm}
\small
\setlength{\tabcolsep}{6pt}
\renewcommand{\arraystretch}{1.12}
\begin{tabular}{ll}
\toprule
Setting & Value \\
\midrule
Patch encoder & CONCH v1.5 \\
Text encoder & Qwen3-Embedding-8B \\
Hidden dimension $d$ & 512 \\
Optimizer & Adam, lr $=1\times10^{-4}$ \\
Batch size & 1 slide \\
Cross-validation & 5-fold \\
Regions $K$ & Dataset/task-specific \\
Evolution steps $Q$ & Dataset/task-specific \\
Neighborhood size & 8 \\
Dropout & 0.3 \\
$\lambda_{\mathrm{div}}$ & 0.1 \\
\bottomrule
\end{tabular}
\vspace{-1mm}
\end{table}

\begin{table*}[!h]
\centering
\caption{Dataset- and task-specific settings of the number of regions $K$ and evolution steps $Q$.}
\label{tab:kt_settings}
\vspace{1mm}
\small
\renewcommand{\arraystretch}{1.12}

\begin{tabular}{@{}ccc@{}}

\begin{minipage}[t]{0.29\textwidth}
\centering
\textbf{Survival Prediction}
\vspace{0.8mm}

\begin{tabular}{lcc}
\toprule
Dataset & $K$ & $Q$ \\
\midrule
LUAD & 6 & 3 \\
GBC  & 4 & 3 \\
BLCA & 8 & 3 \\
BRCA & 6 & 2 \\
\bottomrule
\end{tabular}
\end{minipage}

&

\begin{minipage}[t]{0.36\textwidth}
\centering
\textbf{Subtype Classification}
\vspace{0.8mm}

\begin{tabular}{llcc}
\toprule
Dataset & Level & $K$ & $Q$ \\
\midrule
BRACS   & Fine   & 4 & 2 \\
BRACS   & Coarse & 4 & 2 \\
EBRAINS & Fine   & 4 & 2 \\
EBRAINS & Coarse & 4 & 2 \\
\bottomrule
\end{tabular}
\end{minipage}

&

\begin{minipage}[t]{0.29\textwidth}
\centering
\textbf{Gene Expression Prediction}
\vspace{0.8mm}

\begin{tabular}{lcc}
\toprule
Dataset & $K$ & $Q$ \\
\midrule
LUAD & 6 & 3 \\
BLCA & 8 & 3 \\
BRCA & 8 & 3 \\
--   & -- & -- \\
\bottomrule
\end{tabular}
\end{minipage}

\end{tabular}

\vspace{-1mm}
\end{table*}

\section{Computational Cost Analysis}
\label{app:computational_cost}

We compare the computational cost of TMEvolve with representative WSI-level baselines on LUAD survival prediction. 
All methods in Table~\ref{tab:computational_cost} are evaluated using pre-extracted patch-level WSI features with batch size 1 on a server with eight NVIDIA RTX 4090 GPUs, excluding the offline feature extraction cost. 
We focus on methods that process patch-level features during inference, including MIL baselines, survival-oriented models, and concept-guided WSI-level methods. 
Foundation-model baselines that directly operate on pre-computed slide-level embeddings are not included in this runtime comparison, since their inference mainly applies a lightweight prediction head to already aggregated slide representations.

As shown in Table~\ref{tab:computational_cost}, TMEvolve contains 5.18M trainable parameters and requires 6.48 ms per slide. 
Its parameter size is comparable to structured or concept-guided models such as ConcepPath~\cite{zhao2024aligning} and CITE~\cite{zhang2023text}, while its runtime remains within the millisecond range. 
The additional cost mainly comes from dynamic region assignment, local coordination, and concept-guided boundary interaction over patch-level features. 
Overall, TMEvolve provides a reasonable trade-off between computational cost and explicit microenvironment modeling capacity.

\begin{table*}[!t]
\centering
\caption{Computational cost comparison on LUAD survival prediction. 
Runtime is measured using pre-extracted patch-level WSI features with batch size 1, excluding offline feature extraction.}
\label{tab:computational_cost}
\vspace{1mm}
\small
\setlength{\tabcolsep}{4pt}
\renewcommand{\arraystretch}{1.12}
\resizebox{\textwidth}{!}{
\begin{tabular}{lccccccccccc}
\toprule
Metric 
& ABMIL 
& DSMIL 
& TransMIL 
& RRT-MIL 
& VLSA 
& GECKO 
& ConcepPath 
& ProtoSurv 
& CITE 
& QPMIL 
& TMEvolve \\
\midrule
Trainable Params (M) 
& 1.84 
& 1.91 
& 3.59 
& 3.66 
& 2.77 
& 3.31 
& 4.87 
& 7.03 
& 5.80 
& 2.98 
& 5.18 \\

Time / slide (ms) 
& 0.59 
& 0.94 
& 4.80 
& 4.25 
& 1.51 
& 1.05 
& 1.33 
& 3.94 
& 5.08 
& 1.14 
& 6.48 \\
\bottomrule
\end{tabular}
}
\vspace{-1mm}
\end{table*}

Compared with lightweight MIL baselines, the additional runtime of TMEvolve mainly comes from its iterative evolution module, including region assignment, local coordination, and boundary interaction over patch-level features. 
Nevertheless, the model remains efficient for WSI-level prediction while enabling more explicit microenvironment modeling.

\section{Pathological Concept Construction}
\label{app:concept_construction}

TMEvolve uses language-encoded pathological concepts as semantic priors to guide region-level microenvironment modeling~\cite{huang2023visual,lu2024visual,zhao2024aligning,kapse2025gecko}. 
Instead of relying only on patch-level visual features, we introduce concept descriptions that summarize high-level tissue states related to tumor morphology, immune activity, stromal response, invasion, necrosis, vascular remodeling, and tissue organization~\cite{de2023evolving,quail2013microenvironmental,friedl2011cancer,fridman2012immune,bindea2013spatiotemporal,joyce2015t,kalluri2016biology}. 
These concepts are encoded by the text encoder and used as semantic guidance for region routing and concept-guided boundary interaction.

We organize the concept set into two parts: a shared concept bank and dataset-specific concept banks. 
As shown in Tables~\ref{tab:shared_pathological_concepts} and~\ref{tab:dataset_specific_pathological_concepts}, the shared concept bank contains general histopathological and tumor microenvironment patterns that are broadly applicable across cancer types, while dataset-specific concept banks capture cohort-dependent pathological characteristics. 
The final concept set for each dataset is obtained by combining the shared concepts with its corresponding dataset-specific concepts. 
In the random concept ablation, we replace the language-encoded concept embeddings with random concept probes of the same dimensionality, thereby removing the pathological semantic prior while keeping the parameter budget comparable.

\begin{table*}[!t]
\centering
\caption{Shared pathological concepts used across datasets.}
\label{tab:shared_pathological_concepts}
\vspace{1mm}
\small
\setlength{\tabcolsep}{5pt}
\renewcommand{\arraystretch}{1.18}
\resizebox{\textwidth}{!}{
\begin{tabular}{m{0.16\textwidth}m{0.80\textwidth}}
\toprule
Category & Concept descriptions \\
\midrule
Immune 
& Dense CD8$^+$ T cell infiltration; mature tertiary lymphoid structures; macrophage-rich immune microenvironment with antigen-presenting activity; plasma cell-rich reactive stroma; immune-permissive tumor interface with lymphocyte access. \\

\midrule
Tumor 
& Moderate tumor cellularity with preserved tissue organization; low atypical mitotic activity; well to moderately differentiated tumor glands with limited nuclear pleomorphism; non-solid or less aggressive tumor growth pattern. \\

\midrule
Vascular 
& Limited tumor necrosis with mostly viable tissue; organized microvasculature and stromal remodeling; well-perfused tissue without prominent hemorrhage or ischemia. \\

\midrule
Invasion 
& Focal tumor budding or limited local invasion; partial EMT-like morphology with limited spindle-cell change; mild desmoplastic stromal reaction; CAF-rich stroma with localized stromal activation. \\
\bottomrule
\end{tabular}
}
\vspace{-1mm}
\end{table*}

\begin{table*}[!t]
\centering
\caption{Dataset-specific pathological concepts used in TMEvolve.}
\label{tab:dataset_specific_pathological_concepts}
\vspace{1mm}
\scriptsize
\setlength{\tabcolsep}{3pt}
\renewcommand{\arraystretch}{1.18}
\resizebox{\textwidth}{!}{
\begin{tabular}{m{0.10\textwidth}m{0.82\textwidth}}
\toprule
Dataset & Dataset-specific concept descriptions \\
\midrule
LUAD 
& Lepidic growth pattern; acinar-predominant adenocarcinoma pattern; papillary architecture with preserved glandular differentiation; non-solid lung adenocarcinoma with gland-forming morphology; mucin-producing adenocarcinoma with relatively differentiated morphology; limited or absent tumor spread through air spaces (STAS); alveolar macrophage accumulation with immune-reactive background; collapsed alveoli with mild fibrotic stroma; EGFR-associated lepidic or papillary morphology; KRAS-associated mucinous or solid morphology. \\

\midrule
BRCA 
& Well-differentiated tubular carcinoma morphology; benign fibrocystic changes and organized lobules; low mitotic rate with uniform epithelial cells; invasive ductal carcinoma in stroma; tubule formation and glandular differentiation; high nuclear grade and pleomorphism; DCIS with comedo/cribriform patterns; HER2-enriched high-grade morphology; TNBC-like morphology with pushing borders; HR$^+$ morphology with lower grade; tumor-infiltrating lymphocytes (TILs); stromal fibrosis and desmoplasia; necrotic foci in high-grade carcinoma. \\

\midrule
BLCA 
& Low-grade non-invasive papillary tumors; thickened but organized non-invasive urothelium; absence of muscularis propria invasion; papillary urothelial carcinoma; muscle-invasive urothelial carcinoma; high-grade urothelial carcinoma; CIS-like flat urothelial atypia; squamous differentiation with keratinization; nested or micropapillary variant morphology; inflammation and lymphocytic infiltration; muscle-invasive front with desmoplasia; necrotic and hemorrhagic aggressive regions; basal-like urothelial carcinoma. \\

\midrule
GBC 
& Well-differentiated glands confined to the mucosa; reactive biliary epithelium without true dysplasia; intact muscle layer without invasive nests; adenocarcinoma with irregular glands; biliary-type infiltrative adenocarcinoma; poorly differentiated with solid growth; mucinous differentiation with mucin pools; perineural or lymphovascular invasion; dense desmoplastic reaction; chronic cholecystitis-like background; tumor necrosis and stromal remodeling; muscle or connective tissue invasion; immune infiltration at invasive margin. \\

\midrule
BRACS 
& Normal breast tissue with organized terminal duct lobular units (TDLU); benign lesions such as usual ductal hyperplasia or fibroadenoma; atypical ductal hyperplasia with partial lumen involvement; flat epithelial atypia with uniform round nuclei; ductal carcinoma in situ with intact myoepithelial layer; invasive breast carcinoma with stromal infiltration; well-differentiated morphology with minimal nuclear atypia; high-grade invasive growth with significant pleomorphism and loss of architecture. \\

\midrule
EBRAINS 
& Normal brain parenchyma with preserved neurons and glial cells; low-grade astrocytoma with mild hypercellularity and fibrillary background; oligodendroglioma morphology with classic uniform round nuclei and clear halos; high-grade glioma with prominent microvascular proliferation; glioblastoma morphology featuring pseudopalisading necrosis; dense infiltration of atypical glial cells with marked nuclear pleomorphism; reactive gliosis with evenly distributed gemistocytic astrocytes; tumor margins showing diffuse infiltration into normal white matter. \\
\bottomrule
\end{tabular}
}
\vspace{-1mm}
\end{table*}

\section{Additional Gene Expression Prediction Results}
\label{app:gene_results}

To provide a more comprehensive evaluation of molecular prediction, we report full gene-level results on TCGA-LUAD, TCGA-BLCA, and TCGA-BRCA in Tables~\ref{tab:gene_full_luad_all_models}--\ref{tab:gene_full_brca_all_models}. 
The genes highlighted in the main paper were selected a priori based on clinical relevance and biological interpretability rather than model performance. 
For BLCA, the selected genes include immune and checkpoint-related markers such as CD8A, CTLA4, CXCL9, HAVCR2, PDCD1, and PTPRC, together with MKI67 for proliferation and ERBB2 for clinically relevant receptor signaling. 
For BRCA, we include immune-related genes CD8A, CTLA4, PDCD1, and PTPRC, the proliferation marker MKI67, epithelial and mesenchymal state markers CDH1 and VIM, and EGFR as a receptor signaling gene. 
For LUAD, we focus on immune and inflammatory signaling genes CD274, CD8A, CXCL10, CXCL9, and PDCD1, together with MKI67, EGFR, and MET to reflect proliferation and oncogenic signaling. 
These representative genes assess whether WSI morphology can capture immune activity, proliferation, phenotypic states, and clinically relevant signaling programs across cancer types.

Unlike the main paper, which focuses on these representative genes for compact presentation, the appendix includes all reported genes and all available evaluation metrics. 
We compare TMEvolve with representative MIL baselines, pathology foundation models, and concept-guided gene prediction methods~\cite{pmlr-v80-ilse18a,li2021dual,shao2021transmil,tang2024feature,wang2024pathology,shao2025multiple,ding2025multimodal,zhao2024aligning,kapse2025gecko}. 
Pearson correlation and Spearman correlation measure continuous gene expression prediction, while AUC evaluates the ability to distinguish high- and low-expression groups after median-based binarization. 
All values are multiplied by 100 for compact presentation.

\begingroup
\setlength{\tabcolsep}{1.15pt}
\renewcommand{\arraystretch}{1.02}
\setlength{\LTleft}{0pt}
\setlength{\LTright}{0pt}
\setlength{\LTcapwidth}{\textwidth}
\begin{longtable}{@{\extracolsep{\fill}}
>{\scriptsize}l
>{\scriptsize}l
*{10}{>{\scriptsize}c}
@{}}
\caption{Full gene-level prediction results on TCGA-LUAD. Each cell reports mean$\pm$std over five folds. The best and second-best results for each gene-metric pair are shown in bold and underlined, respectively.}\label{tab:gene_full_luad_all_models}\\
\toprule
Gene & Metric & ABMIL & DSMIL & TransMIL & RRT-MIL & TITAN & CHIEF & Feather & GECKO & ConcepPath & Ours \\
\midrule
\endfirsthead
\caption[]{Full gene-level prediction results on TCGA-LUAD (continued).}\\
\toprule
Gene & Metric & ABMIL & DSMIL & TransMIL & RRT-MIL & TITAN & CHIEF & Feather & GECKO & ConcepPath & Ours \\
\midrule
\endhead
\midrule
\multicolumn{12}{r}{\scriptsize Continued on next page}\\
\endfoot
\bottomrule
\endlastfoot
\multirow{3}{*}{ACTA2} & Pearson & 44.8$\pm$6.6 & 21.2$\pm$11.4 & 36.0$\pm$9.9 & 41.1$\pm$6.9 & \textbf{65.8$\pm$5.3} & \underline{63.2$\pm$6.6} & 54.7$\pm$5.2 & 45.9$\pm$10.0 & 55.1$\pm$8.8 & 55.7$\pm$10.9 \\*
& Spearman & 41.5$\pm$9.7 & 18.2$\pm$13.0 & 32.7$\pm$12.4 & 38.6$\pm$6.7 & \textbf{63.1$\pm$4.6} & \underline{61.9$\pm$7.6} & 52.6$\pm$5.7 & 45.1$\pm$11.3 & 54.2$\pm$9.8 & 53.4$\pm$14.2 \\*
& AUC & 71.1$\pm$3.7 & 57.7$\pm$7.3 & 64.2$\pm$4.8 & 68.3$\pm$3.4 & 78.6$\pm$1.5 & \textbf{79.9$\pm$4.2} & 75.4$\pm$4.4 & 72.3$\pm$6.0 & \underline{78.8$\pm$5.3} & 76.7$\pm$7.3 \\
\multirow{3}{*}{BRAF} & Pearson & 35.7$\pm$8.3 & 26.5$\pm$13.5 & 29.9$\pm$8.9 & 34.6$\pm$10.3 & 42.6$\pm$2.6 & \underline{44.9$\pm$7.9} & 41.4$\pm$4.1 & 44.1$\pm$12.1 & 43.7$\pm$10.9 & \textbf{47.0$\pm$9.8} \\*
& Spearman & 35.7$\pm$5.7 & 24.4$\pm$12.1 & 29.4$\pm$6.9 & 33.0$\pm$6.2 & 38.3$\pm$3.0 & \underline{42.3$\pm$6.0} & 36.2$\pm$3.3 & 40.8$\pm$9.3 & 40.9$\pm$9.0 & \textbf{44.9$\pm$5.7} \\*
& AUC & 67.0$\pm$1.8 & 62.3$\pm$6.2 & 64.2$\pm$3.0 & 66.0$\pm$3.5 & 67.0$\pm$3.3 & \underline{70.0$\pm$4.5} & 65.4$\pm$2.2 & 68.6$\pm$3.8 & 68.7$\pm$2.8 & \textbf{71.8$\pm$2.3} \\
\multirow{3}{*}{CD274} & Pearson & 59.5$\pm$5.4 & 56.1$\pm$5.3 & 56.1$\pm$8.1 & 50.9$\pm$10.0 & 63.4$\pm$6.5 & \textbf{65.6$\pm$4.1} & 58.4$\pm$6.7 & 62.7$\pm$7.0 & \underline{64.4$\pm$6.9} & 63.7$\pm$6.7 \\*
& Spearman & 57.1$\pm$5.5 & 54.2$\pm$5.5 & 51.0$\pm$10.5 & 49.0$\pm$10.3 & 60.2$\pm$5.6 & \textbf{62.9$\pm$4.8} & 57.0$\pm$6.9 & 60.5$\pm$6.9 & \underline{62.2$\pm$5.7} & 60.9$\pm$4.3 \\*
& AUC & 76.9$\pm$3.3 & 77.0$\pm$2.5 & 74.4$\pm$4.5 & 72.4$\pm$5.4 & 79.0$\pm$3.3 & \textbf{81.4$\pm$3.5} & 77.9$\pm$3.9 & 79.6$\pm$3.9 & \underline{80.1$\pm$3.4} & 79.3$\pm$3.1 \\
\multirow{3}{*}{CD4} & Pearson & 57.8$\pm$7.2 & 50.4$\pm$8.9 & 45.2$\pm$12.7 & 48.9$\pm$8.8 & 64.9$\pm$7.1 & \textbf{66.2$\pm$4.1} & 61.8$\pm$3.6 & 61.3$\pm$4.0 & 62.8$\pm$4.4 & \underline{65.5$\pm$3.1} \\*
& Spearman & 53.2$\pm$6.9 & 43.3$\pm$14.6 & 40.2$\pm$10.2 & 45.6$\pm$11.3 & 60.5$\pm$9.1 & \textbf{64.4$\pm$2.1} & 58.8$\pm$2.4 & 61.3$\pm$5.6 & 61.0$\pm$5.1 & \underline{62.3$\pm$6.7} \\*
& AUC & 74.0$\pm$3.0 & 69.4$\pm$7.7 & 67.5$\pm$5.6 & 70.6$\pm$5.4 & \underline{79.6$\pm$3.1} & \textbf{80.6$\pm$1.3} & 76.8$\pm$2.3 & 78.9$\pm$3.9 & 78.7$\pm$3.9 & 79.4$\pm$3.7 \\
\multirow{3}{*}{CD8A} & Pearson & 58.4$\pm$2.3 & 51.0$\pm$4.4 & 51.7$\pm$6.4 & 52.2$\pm$5.2 & 63.6$\pm$5.9 & \underline{65.9$\pm$4.2} & 57.2$\pm$5.6 & 63.4$\pm$5.8 & 64.4$\pm$6.1 & \textbf{68.4$\pm$5.0} \\*
& Spearman & 58.7$\pm$4.8 & 51.4$\pm$4.6 & 51.8$\pm$7.3 & 52.3$\pm$8.5 & 63.5$\pm$8.1 & \underline{66.0$\pm$3.6} & 56.2$\pm$5.5 & 62.6$\pm$4.1 & 63.8$\pm$5.1 & \textbf{66.2$\pm$3.6} \\*
& AUC & 77.3$\pm$2.7 & 73.7$\pm$5.9 & 73.3$\pm$4.7 & 73.9$\pm$4.8 & \underline{80.5$\pm$5.9} & 80.5$\pm$2.5 & 77.5$\pm$2.8 & 78.8$\pm$4.6 & 79.3$\pm$5.3 & \textbf{80.7$\pm$3.3} \\
\multirow{3}{*}{CDH1} & Pearson & 33.3$\pm$10.0 & 30.0$\pm$9.7 & 29.7$\pm$12.5 & 25.7$\pm$11.1 & 35.8$\pm$4.5 & \textbf{46.3$\pm$4.4} & 34.1$\pm$8.3 & 36.9$\pm$10.9 & 39.3$\pm$15.1 & \underline{43.5$\pm$3.3} \\*
& Spearman & 26.1$\pm$6.8 & 23.7$\pm$4.9 & 25.5$\pm$9.7 & 19.3$\pm$7.5 & 35.0$\pm$3.6 & \textbf{40.4$\pm$5.0} & 30.0$\pm$8.7 & 34.1$\pm$7.2 & 34.4$\pm$11.4 & \underline{38.4$\pm$4.1} \\*
& AUC & 61.2$\pm$4.6 & 60.6$\pm$4.7 & 61.6$\pm$4.8 & 57.1$\pm$5.7 & 67.0$\pm$3.3 & \textbf{68.4$\pm$4.9} & 62.5$\pm$3.1 & 66.4$\pm$4.9 & 66.0$\pm$5.3 & \underline{68.3$\pm$6.6} \\
\multirow{3}{*}{CTLA4} & Pearson & 58.9$\pm$4.8 & 52.3$\pm$2.3 & 53.5$\pm$7.5 & 57.0$\pm$7.0 & \textbf{67.9$\pm$1.8} & \underline{67.3$\pm$1.8} & 63.4$\pm$3.5 & 62.5$\pm$4.8 & 65.6$\pm$4.6 & 65.4$\pm$2.4 \\*
& Spearman & 56.3$\pm$5.8 & 50.5$\pm$5.4 & 52.7$\pm$7.1 & 54.1$\pm$5.6 & \underline{66.3$\pm$2.2} & 65.8$\pm$2.2 & 60.1$\pm$5.3 & 61.7$\pm$5.4 & \textbf{66.8$\pm$3.2} & 64.6$\pm$3.3 \\*
& AUC & 74.9$\pm$4.9 & 73.4$\pm$4.2 & 74.2$\pm$4.6 & 73.9$\pm$4.0 & \underline{80.6$\pm$1.6} & 80.2$\pm$1.8 & 78.2$\pm$3.7 & 78.5$\pm$5.2 & \textbf{81.6$\pm$3.8} & 79.2$\pm$4.8 \\
\multirow{3}{*}{CXCL10} & Pearson & 60.6$\pm$4.4 & 56.1$\pm$8.3 & 56.9$\pm$6.0 & 53.1$\pm$6.9 & 61.9$\pm$5.3 & 64.8$\pm$5.1 & 59.7$\pm$5.1 & 65.3$\pm$4.2 & \textbf{66.6$\pm$7.4} & \underline{65.9$\pm$4.5} \\*
& Spearman & 60.5$\pm$6.0 & 55.3$\pm$8.1 & 57.0$\pm$6.5 & 52.2$\pm$8.9 & 62.3$\pm$4.6 & 65.0$\pm$5.1 & 59.2$\pm$4.4 & 63.9$\pm$4.6 & \textbf{67.3$\pm$10.5} & \underline{65.5$\pm$5.7} \\*
& AUC & 79.3$\pm$4.7 & 77.5$\pm$6.0 & 77.2$\pm$3.9 & 75.6$\pm$4.4 & 81.3$\pm$3.1 & 81.8$\pm$1.8 & 78.4$\pm$3.8 & 82.5$\pm$2.5 & \textbf{83.3$\pm$7.1} & \underline{83.1$\pm$3.2} \\
\multirow{3}{*}{CXCL9} & Pearson & 60.7$\pm$5.4 & 53.1$\pm$4.4 & 49.2$\pm$12.7 & 53.5$\pm$10.4 & 63.2$\pm$4.3 & 62.1$\pm$4.5 & 56.5$\pm$4.7 & 61.9$\pm$5.3 & \underline{66.1$\pm$3.7} & \textbf{66.3$\pm$3.3} \\*
& Spearman & 59.5$\pm$5.0 & 51.9$\pm$5.9 & 49.7$\pm$12.6 & 54.6$\pm$11.3 & 61.4$\pm$5.9 & 63.0$\pm$4.5 & 56.5$\pm$5.7 & 61.9$\pm$5.5 & \textbf{65.7$\pm$6.0} & \underline{64.7$\pm$5.7} \\*
& AUC & 79.0$\pm$3.7 & 77.5$\pm$4.1 & 74.4$\pm$8.0 & 77.8$\pm$7.4 & 81.7$\pm$5.8 & \underline{82.9$\pm$3.8} & 79.7$\pm$3.2 & 82.3$\pm$3.9 & \textbf{83.6$\pm$4.8} & 82.8$\pm$4.7 \\
\multirow{3}{*}{EGFR} & Pearson & 32.4$\pm$6.8 & 20.5$\pm$6.4 & 26.8$\pm$8.3 & 36.2$\pm$5.7 & 40.3$\pm$9.0 & 38.7$\pm$11.0 & 35.2$\pm$10.4 & 41.1$\pm$9.6 & \underline{43.7$\pm$10.0} & \textbf{44.3$\pm$8.6} \\*
& Spearman & 35.9$\pm$9.6 & 26.8$\pm$5.8 & 26.2$\pm$10.7 & 38.0$\pm$7.3 & 41.8$\pm$7.8 & \underline{42.5$\pm$7.9} & 35.3$\pm$6.9 & 41.4$\pm$10.9 & 42.3$\pm$9.4 & \textbf{42.8$\pm$9.1} \\*
& AUC & 68.3$\pm$3.4 & 63.5$\pm$6.6 & 63.1$\pm$7.3 & 67.7$\pm$3.4 & \underline{71.1$\pm$4.7} & \textbf{71.6$\pm$3.4} & 66.8$\pm$3.1 & 71.0$\pm$4.7 & 71.1$\pm$4.4 & 69.7$\pm$3.9 \\
\multirow{3}{*}{ERBB2} & Pearson & 35.9$\pm$11.8 & 37.7$\pm$9.7 & 28.6$\pm$7.7 & 23.7$\pm$9.9 & 41.3$\pm$4.1 & 39.4$\pm$6.5 & 43.4$\pm$7.7 & 41.9$\pm$6.0 & \textbf{49.0$\pm$12.8} & \underline{48.5$\pm$7.8} \\*
& Spearman & 31.1$\pm$8.8 & 34.6$\pm$5.8 & 24.7$\pm$4.1 & 20.2$\pm$7.9 & 35.8$\pm$4.4 & 33.8$\pm$4.7 & 36.7$\pm$8.0 & 36.8$\pm$6.3 & \textbf{42.8$\pm$10.2} & \underline{41.7$\pm$4.7} \\*
& AUC & 65.3$\pm$5.0 & 65.1$\pm$4.9 & 63.0$\pm$2.1 & 58.5$\pm$2.1 & 66.8$\pm$3.9 & 66.2$\pm$4.2 & 67.3$\pm$4.3 & 67.0$\pm$5.2 & \textbf{69.8$\pm$5.5} & \underline{69.0$\pm$2.2} \\
\multirow{3}{*}{FOXP3} & Pearson & 50.4$\pm$9.4 & 48.1$\pm$9.1 & 43.8$\pm$9.2 & 45.8$\pm$8.3 & \textbf{58.8$\pm$5.6} & 57.1$\pm$4.1 & 53.5$\pm$6.2 & 56.4$\pm$7.6 & \underline{58.7$\pm$7.1} & 58.5$\pm$6.4 \\*
& Spearman & 46.6$\pm$8.2 & 43.4$\pm$12.5 & 39.3$\pm$8.1 & 40.7$\pm$9.3 & \textbf{57.5$\pm$7.1} & 55.9$\pm$4.9 & 53.0$\pm$5.1 & 55.4$\pm$10.2 & \underline{57.5$\pm$10.6} & 57.2$\pm$9.2 \\*
& AUC & 72.0$\pm$5.7 & 71.3$\pm$5.8 & 67.2$\pm$3.9 & 68.3$\pm$5.7 & 77.9$\pm$3.5 & 76.8$\pm$3.1 & 75.9$\pm$2.8 & 77.1$\pm$4.2 & \textbf{78.5$\pm$5.1} & \underline{78.5$\pm$4.7} \\
\multirow{3}{*}{HAVCR2} & Pearson & 61.9$\pm$4.9 & 56.5$\pm$5.9 & 58.3$\pm$5.7 & 59.6$\pm$3.3 & 66.1$\pm$4.2 & 67.5$\pm$3.8 & 63.5$\pm$2.4 & 64.0$\pm$5.0 & \underline{67.9$\pm$7.0} & \textbf{68.1$\pm$4.7} \\*
& Spearman & 56.7$\pm$7.8 & 53.5$\pm$7.8 & 55.1$\pm$5.6 & 55.2$\pm$3.6 & 60.4$\pm$3.6 & \underline{64.9$\pm$4.4} & 59.3$\pm$2.5 & 62.4$\pm$5.9 & \textbf{65.5$\pm$3.1} & 64.6$\pm$3.9 \\*
& AUC & 75.6$\pm$5.6 & 75.1$\pm$4.9 & 75.9$\pm$3.4 & 74.9$\pm$4.2 & 77.3$\pm$2.6 & 79.1$\pm$2.9 & 75.5$\pm$4.1 & 79.0$\pm$6.6 & \textbf{80.9$\pm$3.3} & \underline{80.6$\pm$4.2} \\
\multirow{3}{*}{HIF1A} & Pearson & 51.0$\pm$6.1 & 48.9$\pm$11.3 & 47.0$\pm$6.7 & 38.7$\pm$13.9 & 52.5$\pm$6.5 & 49.1$\pm$4.6 & 46.6$\pm$5.1 & 53.1$\pm$7.6 & \underline{55.8$\pm$5.9} & \textbf{58.4$\pm$4.3} \\*
& Spearman & 50.0$\pm$7.4 & 47.8$\pm$14.0 & 46.4$\pm$10.0 & 37.8$\pm$14.1 & 50.5$\pm$8.0 & 47.6$\pm$7.1 & 45.5$\pm$7.2 & 50.8$\pm$8.7 & \underline{52.4$\pm$8.3} & \textbf{55.2$\pm$5.9} \\*
& AUC & 77.1$\pm$5.5 & 75.4$\pm$8.5 & 74.5$\pm$5.3 & 69.7$\pm$8.5 & 75.8$\pm$5.8 & 74.4$\pm$3.4 & 73.4$\pm$5.3 & 77.6$\pm$5.5 & \underline{77.7$\pm$5.6} & \textbf{78.5$\pm$3.3} \\
\multirow{3}{*}{KEAP1} & Pearson & 4.3$\pm$13.6 & -1.8$\pm$12.6 & 13.4$\pm$12.0 & 4.4$\pm$6.7 & 24.0$\pm$11.1 & \textbf{25.3$\pm$14.3} & 18.6$\pm$8.9 & 20.8$\pm$8.7 & 22.0$\pm$5.4 & \underline{24.5$\pm$6.1} \\*
& Spearman & 8.7$\pm$13.9 & 0.7$\pm$9.8 & 10.2$\pm$6.8 & 3.6$\pm$5.7 & 22.1$\pm$12.0 & \textbf{26.7$\pm$11.2} & 16.7$\pm$9.7 & 21.0$\pm$11.5 & 23.2$\pm$10.5 & \underline{26.4$\pm$5.7} \\*
& AUC & 55.9$\pm$6.7 & 52.4$\pm$5.4 & 53.0$\pm$2.7 & 51.6$\pm$4.4 & 60.6$\pm$7.4 & \underline{63.0$\pm$5.9} & 58.0$\pm$3.8 & 61.1$\pm$4.8 & 61.9$\pm$4.8 & \textbf{63.2$\pm$1.9} \\
\multirow{3}{*}{KRAS} & Pearson & 17.4$\pm$12.4 & 13.6$\pm$7.8 & 14.5$\pm$11.0 & 9.0$\pm$13.1 & \textbf{31.3$\pm$1.0} & 30.1$\pm$6.7 & 27.6$\pm$4.5 & 26.0$\pm$10.0 & 29.9$\pm$11.5 & \underline{31.1$\pm$13.1} \\*
& Spearman & 15.2$\pm$10.3 & 13.1$\pm$7.2 & 9.7$\pm$10.4 & 7.1$\pm$14.1 & \textbf{31.5$\pm$5.6} & \underline{30.2$\pm$3.7} & 25.3$\pm$4.4 & 24.5$\pm$7.9 & 22.4$\pm$11.2 & 24.8$\pm$13.3 \\*
& AUC & 56.5$\pm$2.5 & 57.9$\pm$6.5 & 54.9$\pm$4.1 & 52.7$\pm$9.1 & \textbf{63.2$\pm$4.6} & \underline{63.1$\pm$2.8} & 61.7$\pm$3.1 & 61.8$\pm$5.0 & 60.2$\pm$5.3 & 62.8$\pm$7.8 \\
\multirow{3}{*}{LAG3} & Pearson & 57.8$\pm$5.7 & 53.2$\pm$6.5 & 52.0$\pm$8.8 & 48.0$\pm$8.5 & 62.0$\pm$5.8 & 62.5$\pm$5.2 & 55.3$\pm$6.6 & 62.8$\pm$4.2 & \underline{65.0$\pm$7.8} & \textbf{65.0$\pm$5.5} \\*
& Spearman & 56.4$\pm$7.3 & 50.8$\pm$8.5 & 49.7$\pm$8.2 & 47.3$\pm$9.8 & 60.6$\pm$5.9 & 62.4$\pm$6.2 & 53.9$\pm$6.9 & 61.6$\pm$6.6 & \underline{64.1$\pm$10.2} & \textbf{64.5$\pm$6.5} \\*
& AUC & 76.7$\pm$4.9 & 74.1$\pm$6.3 & 73.2$\pm$4.7 & 72.5$\pm$7.2 & 80.8$\pm$4.4 & 80.8$\pm$3.9 & 74.9$\pm$3.7 & \underline{81.6$\pm$4.9} & 81.1$\pm$7.2 & \textbf{81.7$\pm$5.1} \\
\multirow{3}{*}{MET} & Pearson & 49.8$\pm$2.2 & 44.5$\pm$8.0 & 43.1$\pm$6.7 & 46.0$\pm$7.1 & 53.8$\pm$5.1 & 49.5$\pm$4.4 & 42.9$\pm$7.2 & \textbf{59.7$\pm$4.4} & \underline{58.2$\pm$2.3} & 57.2$\pm$5.6 \\*
& Spearman & 48.6$\pm$4.1 & 42.3$\pm$9.4 & 41.0$\pm$8.1 & 41.4$\pm$7.2 & 50.4$\pm$4.6 & 47.7$\pm$5.6 & 35.7$\pm$8.8 & \textbf{57.1$\pm$6.1} & \underline{54.4$\pm$5.8} & 54.0$\pm$7.6 \\*
& AUC & 73.2$\pm$2.8 & 71.2$\pm$6.3 & 68.7$\pm$5.1 & 68.1$\pm$2.9 & 74.7$\pm$3.5 & 72.1$\pm$2.7 & 65.4$\pm$4.9 & \textbf{77.7$\pm$3.0} & \underline{76.5$\pm$4.3} & 75.4$\pm$3.3 \\
\multirow{3}{*}{MKI67} & Pearson & 63.6$\pm$1.8 & 59.9$\pm$3.3 & 56.3$\pm$3.2 & 55.9$\pm$3.0 & \textbf{72.3$\pm$2.7} & \underline{71.6$\pm$2.5} & 70.2$\pm$4.7 & 65.1$\pm$4.7 & 65.5$\pm$6.3 & 67.0$\pm$4.0 \\*
& Spearman & 63.3$\pm$2.2 & 60.6$\pm$4.6 & 54.4$\pm$4.1 & 54.9$\pm$2.3 & \textbf{72.2$\pm$2.7} & \underline{71.5$\pm$3.2} & 70.2$\pm$4.6 & 66.0$\pm$5.1 & 65.7$\pm$5.5 & 67.1$\pm$3.7 \\*
& AUC & 81.8$\pm$2.7 & 80.1$\pm$4.9 & 76.0$\pm$3.0 & 78.0$\pm$2.7 & \underline{87.0$\pm$3.0} & 85.9$\pm$2.3 & \textbf{87.2$\pm$4.7} & 82.6$\pm$4.8 & 82.3$\pm$4.7 & 83.1$\pm$4.5 \\
\multirow{3}{*}{PDCD1} & Pearson & 61.9$\pm$6.1 & 58.2$\pm$7.2 & 54.0$\pm$10.2 & 59.5$\pm$3.5 & 65.6$\pm$4.9 & 63.3$\pm$3.5 & 60.1$\pm$4.3 & \underline{65.8$\pm$4.5} & \textbf{68.9$\pm$7.1} & 65.3$\pm$6.7 \\*
& Spearman & 61.5$\pm$7.4 & 54.6$\pm$11.2 & 53.3$\pm$11.3 & 55.7$\pm$6.3 & \underline{64.7$\pm$6.4} & 62.0$\pm$4.6 & 58.1$\pm$4.1 & 64.6$\pm$5.4 & \textbf{67.6$\pm$8.6} & 63.8$\pm$8.4 \\*
& AUC & 80.0$\pm$5.1 & 76.4$\pm$5.4 & 75.6$\pm$6.0 & 76.8$\pm$6.0 & 82.0$\pm$4.6 & 80.0$\pm$4.1 & 77.5$\pm$2.8 & \textbf{82.7$\pm$2.3} & \underline{82.7$\pm$4.7} & 81.8$\pm$5.1 \\
\multirow{3}{*}{PTPRC} & Pearson & 57.7$\pm$6.9 & 49.3$\pm$6.0 & 49.2$\pm$15.8 & 51.6$\pm$13.8 & \underline{65.5$\pm$7.2} & \textbf{66.4$\pm$4.0} & 63.3$\pm$3.3 & 59.7$\pm$1.7 & 62.1$\pm$4.1 & 64.1$\pm$1.5 \\*
& Spearman & 57.1$\pm$6.5 & 45.7$\pm$10.3 & 49.7$\pm$12.9 & 50.7$\pm$14.2 & \underline{63.4$\pm$5.8} & \textbf{66.1$\pm$2.8} & 62.6$\pm$3.3 & 59.6$\pm$2.6 & 62.6$\pm$4.1 & 63.0$\pm$2.9 \\*
& AUC & 79.0$\pm$3.8 & 72.2$\pm$6.6 & 74.5$\pm$5.6 & 75.0$\pm$7.7 & \textbf{80.8$\pm$5.6} & 80.4$\pm$4.1 & 79.4$\pm$3.8 & 80.0$\pm$2.5 & \underline{80.7$\pm$2.4} & 80.6$\pm$3.4 \\
\multirow{3}{*}{STK11} & Pearson & 25.7$\pm$8.2 & 22.9$\pm$10.4 & 23.3$\pm$5.3 & 17.4$\pm$15.1 & \underline{39.9$\pm$7.8} & 37.2$\pm$6.2 & 37.0$\pm$3.9 & 38.1$\pm$5.0 & 39.7$\pm$6.4 & \textbf{42.0$\pm$9.3} \\*
& Spearman & 22.5$\pm$9.1 & 18.9$\pm$8.4 & 19.2$\pm$6.8 & 14.3$\pm$15.7 & \textbf{36.7$\pm$8.7} & 35.2$\pm$8.6 & 33.0$\pm$8.7 & 33.2$\pm$6.5 & 35.6$\pm$5.2 & \underline{35.9$\pm$10.4} \\*
& AUC & 58.8$\pm$4.8 & 58.7$\pm$6.3 & 58.6$\pm$5.3 & 54.8$\pm$9.9 & \textbf{68.3$\pm$7.1} & \underline{67.8$\pm$7.3} & 64.8$\pm$6.5 & 65.8$\pm$6.5 & 66.7$\pm$4.3 & 66.0$\pm$5.6 \\
\multirow{3}{*}{TP53} & Pearson & 21.6$\pm$10.4 & 21.2$\pm$12.0 & 24.8$\pm$11.4 & 16.6$\pm$13.1 & 17.1$\pm$10.1 & \underline{27.8$\pm$7.7} & 19.5$\pm$13.7 & 25.8$\pm$6.4 & \textbf{30.7$\pm$8.7} & 23.4$\pm$8.8 \\*
& Spearman & 19.4$\pm$11.5 & 18.6$\pm$13.1 & 22.4$\pm$8.5 & 15.7$\pm$14.1 & 19.1$\pm$9.4 & \textbf{29.6$\pm$9.3} & 19.2$\pm$12.8 & 23.4$\pm$7.8 & \underline{27.0$\pm$11.7} & 23.0$\pm$12.1 \\*
& AUC & 61.3$\pm$6.2 & 60.8$\pm$7.6 & 62.5$\pm$4.7 & 59.9$\pm$5.7 & 61.8$\pm$5.1 & \textbf{67.4$\pm$6.3} & 61.4$\pm$7.2 & 63.6$\pm$2.9 & \underline{64.3$\pm$4.9} & 63.3$\pm$6.8 \\
\multirow{3}{*}{VIM} & Pearson & 50.2$\pm$9.5 & 38.6$\pm$6.6 & 41.4$\pm$3.9 & 40.7$\pm$11.7 & 56.8$\pm$5.1 & 57.9$\pm$4.3 & 47.8$\pm$8.0 & 56.5$\pm$6.4 & \underline{59.8$\pm$7.3} & \textbf{61.5$\pm$6.0} \\*
& Spearman & 48.2$\pm$11.6 & 34.6$\pm$7.3 & 41.1$\pm$6.6 & 38.3$\pm$10.5 & 55.2$\pm$5.8 & 57.1$\pm$5.7 & 47.8$\pm$5.5 & 55.2$\pm$6.9 & \underline{58.5$\pm$7.2} & \textbf{59.0$\pm$9.0} \\*
& AUC & 73.0$\pm$7.4 & 64.5$\pm$3.9 & 68.5$\pm$4.7 & 66.0$\pm$4.7 & 76.4$\pm$2.9 & 77.7$\pm$3.4 & 72.6$\pm$3.7 & 76.4$\pm$5.0 & \underline{78.4$\pm$5.3} & \textbf{79.0$\pm$6.7} \\
\end{longtable}
\endgroup

\begingroup
\setlength{\tabcolsep}{1.15pt}
\renewcommand{\arraystretch}{1.02}
\setlength{\LTleft}{0pt}
\setlength{\LTright}{0pt}
\setlength{\LTcapwidth}{\textwidth}
\begin{longtable}{@{\extracolsep{\fill}}
>{\scriptsize}l
>{\scriptsize}l
*{10}{>{\scriptsize}c}
@{}}
\caption{Full gene-level prediction results on TCGA-BLCA. Each cell reports mean$\pm$std over five folds. The best and second-best results for each gene-metric pair are shown in bold and underlined, respectively.}\label{tab:gene_full_blca_all_models}\\
\toprule
Gene & Metric & ABMIL & DSMIL & TransMIL & RRT-MIL & TITAN & CHIEF & Feather & GECKO & ConcepPath & Ours \\
\midrule
\endfirsthead
\caption[]{Full gene-level prediction results on TCGA-BLCA (continued).}\\
\toprule
Gene & Metric & ABMIL & DSMIL & TransMIL & RRT-MIL & TITAN & CHIEF & Feather & GECKO & ConcepPath& Ours \\
\midrule
\endhead
\midrule
\multicolumn{12}{r}{\scriptsize Continued on next page}\\
\endfoot
\bottomrule
\endlastfoot
\multirow{3}{*}{ACTA2} & Pearson & 59.0$\pm$6.5 & 50.1$\pm$3.0 & 52.1$\pm$6.9 & 57.2$\pm$10.1 & 46.0$\pm$14.3 & \underline{72.1$\pm$2.3} & 67.5$\pm$2.6 & 71.2$\pm$1.6 & \textbf{73.5$\pm$2.2} & 71.7$\pm$1.4 \\*
& Spearman & 59.2$\pm$8.0 & 49.4$\pm$4.4 & 53.4$\pm$5.2 & 57.4$\pm$11.1 & 45.1$\pm$15.8 & \underline{72.4$\pm$3.8} & 67.8$\pm$2.6 & 72.0$\pm$2.4 & \textbf{72.6$\pm$3.9} & 71.9$\pm$1.9 \\*
& AUC & 77.6$\pm$8.1 & 73.0$\pm$3.2 & 74.5$\pm$5.1 & 76.7$\pm$10.4 & 72.9$\pm$12.6 & 84.8$\pm$4.7 & 83.1$\pm$3.3 & 84.9$\pm$3.9 & \textbf{85.1$\pm$3.2} & \underline{85.1$\pm$3.0} \\
\multirow{3}{*}{BRAF} & Pearson & 28.8$\pm$14.4 & 33.8$\pm$16.3 & 32.7$\pm$13.3 & 26.2$\pm$11.4 & 24.9$\pm$18.2 & 45.6$\pm$5.7 & 43.6$\pm$9.0 & \underline{47.9$\pm$6.3} & 47.8$\pm$7.2 & \textbf{49.2$\pm$6.2} \\*
& Spearman & 26.7$\pm$8.6 & 27.1$\pm$13.0 & 28.7$\pm$11.3 & 22.1$\pm$11.4 & 23.0$\pm$16.9 & 41.8$\pm$6.0 & 38.0$\pm$10.1 & 43.3$\pm$6.1 & \underline{43.5$\pm$5.6} & \textbf{46.3$\pm$5.4} \\*
& AUC & 64.5$\pm$5.0 & 60.2$\pm$5.5 & 63.2$\pm$6.9 & 61.5$\pm$3.8 & 60.1$\pm$9.3 & 69.6$\pm$4.9 & 68.5$\pm$5.1 & 71.8$\pm$3.6 & \underline{72.3$\pm$2.1} & \textbf{73.3$\pm$3.9} \\
\multirow{3}{*}{CD274} & Pearson & 60.3$\pm$9.3 & 53.7$\pm$6.3 & 54.1$\pm$5.4 & 41.2$\pm$13.7 & 37.5$\pm$14.5 & 64.4$\pm$6.3 & 57.9$\pm$5.9 & 64.9$\pm$7.4 & \underline{66.0$\pm$7.4} & \textbf{66.6$\pm$4.8} \\*
& Spearman & 61.4$\pm$8.8 & 52.2$\pm$9.3 & 54.8$\pm$5.9 & 40.0$\pm$17.9 & 37.0$\pm$19.0 & 64.0$\pm$4.7 & 56.3$\pm$4.9 & 64.2$\pm$8.0 & \underline{65.2$\pm$8.2} & \textbf{66.0$\pm$5.4} \\*
& AUC & 81.8$\pm$6.2 & 77.2$\pm$5.5 & 78.7$\pm$3.4 & 70.4$\pm$8.9 & 65.5$\pm$12.4 & 82.3$\pm$3.8 & 78.9$\pm$2.0 & 82.1$\pm$6.0 & \underline{83.1$\pm$6.5} & \textbf{83.4$\pm$4.6} \\
\multirow{3}{*}{CD4} & Pearson & 60.8$\pm$5.9 & 52.2$\pm$6.9 & 53.7$\pm$8.1 & 54.1$\pm$5.4 & 44.1$\pm$12.8 & 62.8$\pm$1.4 & 62.3$\pm$6.4 & 69.7$\pm$5.8 & \textbf{71.0$\pm$8.2} & \underline{70.9$\pm$6.6} \\*
& Spearman & 61.4$\pm$3.6 & 50.3$\pm$10.2 & 54.7$\pm$6.2 & 55.6$\pm$7.6 & 45.4$\pm$12.5 & 62.6$\pm$2.2 & 62.3$\pm$7.5 & 69.6$\pm$6.8 & \underline{70.5$\pm$8.3} & \textbf{70.6$\pm$7.1} \\*
& AUC & 79.8$\pm$1.8 & 75.8$\pm$6.2 & 78.4$\pm$1.7 & 78.1$\pm$3.0 & 73.0$\pm$7.7 & 81.7$\pm$2.6 & 82.1$\pm$3.8 & 85.3$\pm$4.7 & \textbf{86.0$\pm$4.5} & \underline{85.8$\pm$4.3} \\
\multirow{3}{*}{CD8A} & Pearson & 65.2$\pm$8.7 & 57.7$\pm$8.2 & 58.5$\pm$8.7 & 55.3$\pm$7.7 & 56.6$\pm$4.2 & 66.5$\pm$6.4 & 63.3$\pm$5.4 & \underline{71.7$\pm$3.2} & 71.2$\pm$3.3 & \textbf{73.5$\pm$5.6} \\*
& Spearman & 62.4$\pm$8.9 & 53.2$\pm$9.4 & 54.5$\pm$9.9 & 52.0$\pm$8.5 & 54.0$\pm$1.9 & 63.4$\pm$5.2 & 59.8$\pm$5.1 & \underline{70.4$\pm$2.1} & 69.0$\pm$3.2 & \textbf{71.6$\pm$4.6} \\*
& AUC & 79.7$\pm$4.0 & 75.1$\pm$4.4 & 74.4$\pm$5.2 & 73.4$\pm$5.7 & 77.7$\pm$3.5 & 80.1$\pm$4.3 & 77.7$\pm$3.4 & \textbf{83.3$\pm$2.4} & 81.3$\pm$1.6 & \underline{83.0$\pm$3.2} \\
\multirow{3}{*}{CDH1} & Pearson & 51.9$\pm$6.9 & 33.2$\pm$9.5 & 32.3$\pm$17.2 & 41.4$\pm$17.5 & 11.0$\pm$20.5 & 52.6$\pm$4.5 & 50.2$\pm$11.8 & \underline{58.0$\pm$8.4} & 54.4$\pm$7.6 & \textbf{61.0$\pm$6.9} \\*
& Spearman & 32.4$\pm$14.9 & 25.4$\pm$16.8 & 23.1$\pm$10.5 & 21.8$\pm$18.6 & 4.1$\pm$17.6 & 38.5$\pm$10.6 & 38.4$\pm$7.2 & \textbf{41.9$\pm$11.0} & \underline{39.5$\pm$10.8} & 39.5$\pm$12.3 \\*
& AUC & 63.3$\pm$9.9 & 62.5$\pm$11.7 & 60.4$\pm$6.0 & 59.5$\pm$10.3 & 49.2$\pm$9.1 & \textbf{68.4$\pm$6.8} & 66.8$\pm$4.7 & \underline{67.8$\pm$7.4} & 67.6$\pm$8.1 & 65.8$\pm$6.7 \\
\multirow{3}{*}{CTLA4} & Pearson & 65.1$\pm$4.1 & 61.6$\pm$6.4 & 60.4$\pm$5.2 & 57.7$\pm$6.8 & 54.8$\pm$3.2 & 66.0$\pm$7.1 & 66.6$\pm$6.1 & \underline{70.1$\pm$5.3} & 68.9$\pm$2.4 & \textbf{72.7$\pm$4.9} \\*
& Spearman & 65.6$\pm$4.6 & 60.5$\pm$8.4 & 58.5$\pm$5.7 & 58.1$\pm$6.7 & 55.4$\pm$6.9 & 63.8$\pm$6.4 & 65.4$\pm$8.5 & \underline{68.6$\pm$6.2} & 66.7$\pm$3.4 & \textbf{70.6$\pm$6.6} \\*
& AUC & 82.1$\pm$2.6 & 79.1$\pm$3.3 & 77.1$\pm$7.0 & 78.7$\pm$5.6 & 78.2$\pm$4.8 & 80.1$\pm$5.0 & 81.1$\pm$6.2 & \textbf{83.5$\pm$5.9} & 82.2$\pm$4.4 & \underline{83.5$\pm$5.9} \\
\multirow{3}{*}{CXCL10} & Pearson & 63.8$\pm$2.8 & 59.8$\pm$6.3 & 61.4$\pm$2.1 & 55.6$\pm$5.1 & 44.8$\pm$18.4 & 66.1$\pm$7.1 & 65.5$\pm$5.7 & \underline{70.3$\pm$3.2} & 69.3$\pm$3.3 & \textbf{71.6$\pm$4.8} \\*
& Spearman & 62.5$\pm$4.4 & 56.2$\pm$9.2 & 59.6$\pm$4.0 & 53.3$\pm$5.8 & 44.5$\pm$23.8 & 65.4$\pm$8.3 & 64.8$\pm$7.7 & \underline{70.3$\pm$5.0} & 68.8$\pm$4.1 & \textbf{71.1$\pm$5.6} \\*
& AUC & 80.8$\pm$5.1 & 75.3$\pm$6.0 & 80.1$\pm$3.2 & 76.0$\pm$3.2 & 68.4$\pm$12.4 & 82.6$\pm$4.9 & 80.3$\pm$5.2 & \textbf{85.2$\pm$5.0} & 83.2$\pm$4.4 & \underline{85.1$\pm$4.8} \\
\multirow{3}{*}{CXCL9} & Pearson & 62.9$\pm$3.4 & 55.8$\pm$8.0 & 60.6$\pm$2.5 & 56.3$\pm$9.8 & 58.0$\pm$7.8 & 66.5$\pm$7.9 & 65.1$\pm$6.4 & 72.4$\pm$2.4 & \underline{72.4$\pm$1.6} & \textbf{74.9$\pm$1.5} \\*
& Spearman & 62.5$\pm$3.1 & 54.1$\pm$9.3 & 59.5$\pm$2.2 & 54.4$\pm$9.7 & 53.2$\pm$10.2 & 66.2$\pm$7.1 & 64.3$\pm$8.3 & \underline{72.6$\pm$3.8} & 71.6$\pm$1.7 & \textbf{74.0$\pm$2.5} \\*
& AUC & 80.1$\pm$3.6 & 75.1$\pm$4.4 & 76.6$\pm$4.4 & 75.4$\pm$5.4 & 74.1$\pm$8.2 & 83.3$\pm$3.5 & 82.3$\pm$6.0 & 85.6$\pm$4.8 & \underline{85.7$\pm$2.5} & \textbf{86.8$\pm$2.6} \\
\multirow{3}{*}{EGFR} & Pearson & 40.9$\pm$8.0 & 36.8$\pm$7.1 & 33.7$\pm$14.4 & 35.9$\pm$4.9 & \textbf{54.2$\pm$17.8} & 46.1$\pm$9.0 & 40.5$\pm$6.3 & 45.5$\pm$8.1 & 43.5$\pm$9.7 & \underline{46.1$\pm$6.2} \\*
& Spearman & 44.7$\pm$7.3 & 41.5$\pm$4.7 & 33.2$\pm$14.1 & 37.9$\pm$6.6 & \textbf{51.8$\pm$18.0} & \underline{49.0$\pm$8.6} & 41.5$\pm$6.1 & 47.3$\pm$6.2 & 45.0$\pm$6.6 & 48.1$\pm$5.7 \\*
& AUC & 72.6$\pm$2.7 & 70.5$\pm$1.6 & 67.0$\pm$5.3 & 68.3$\pm$3.3 & \textbf{76.9$\pm$11.8} & \underline{74.5$\pm$3.7} & 71.2$\pm$2.6 & 74.3$\pm$4.0 & 73.0$\pm$1.1 & 74.0$\pm$3.7 \\
\multirow{3}{*}{ERBB2} & Pearson & 55.1$\pm$2.2 & 50.8$\pm$8.0 & 41.4$\pm$8.3 & 41.8$\pm$7.9 & 48.4$\pm$18.2 & 49.4$\pm$5.4 & 52.5$\pm$7.7 & \underline{60.0$\pm$4.7} & 59.8$\pm$7.7 & \textbf{62.0$\pm$6.2} \\*
& Spearman & 53.8$\pm$7.9 & 52.2$\pm$8.0 & 42.3$\pm$11.7 & 46.0$\pm$9.8 & 48.3$\pm$17.3 & 55.3$\pm$3.0 & 54.7$\pm$6.5 & \underline{61.7$\pm$5.8} & 61.5$\pm$5.5 & \textbf{62.9$\pm$7.7} \\*
& AUC & 75.8$\pm$8.1 & 77.5$\pm$4.4 & 70.8$\pm$7.0 & 74.4$\pm$6.3 & 71.7$\pm$10.8 & 78.0$\pm$1.4 & 76.6$\pm$4.3 & \underline{80.4$\pm$4.0} & \textbf{81.1$\pm$3.0} & 80.4$\pm$4.2 \\
\multirow{3}{*}{FOXP3} & Pearson & 54.2$\pm$7.3 & 55.8$\pm$7.3 & 54.7$\pm$6.8 & 49.0$\pm$8.0 & 43.7$\pm$17.2 & 59.5$\pm$3.3 & 58.3$\pm$9.0 & 62.7$\pm$3.5 & \underline{63.5$\pm$3.0} & \textbf{64.1$\pm$4.8} \\*
& Spearman & 54.0$\pm$9.0 & 55.6$\pm$8.4 & 54.0$\pm$7.1 & 48.5$\pm$9.6 & 41.7$\pm$23.1 & 58.9$\pm$4.6 & 57.4$\pm$9.6 & 62.4$\pm$3.4 & \underline{63.4$\pm$3.4} & \textbf{63.9$\pm$5.0} \\*
& AUC & 77.7$\pm$4.2 & 78.3$\pm$3.4 & 79.0$\pm$5.0 & 75.2$\pm$5.8 & 72.6$\pm$17.5 & 79.6$\pm$2.8 & 78.7$\pm$5.0 & 80.9$\pm$3.3 & \textbf{81.4$\pm$2.6} & \underline{81.0$\pm$2.6} \\
\multirow{3}{*}{HAVCR2} & Pearson & 71.8$\pm$2.4 & 62.6$\pm$5.2 & 65.7$\pm$7.9 & 69.8$\pm$5.4 & 59.4$\pm$13.8 & 76.8$\pm$3.4 & 73.0$\pm$4.7 & \underline{79.4$\pm$3.6} & 77.8$\pm$3.1 & \textbf{80.2$\pm$3.1} \\*
& Spearman & 72.8$\pm$1.0 & 60.7$\pm$5.3 & 66.6$\pm$7.2 & 67.9$\pm$6.0 & 54.7$\pm$14.7 & 77.4$\pm$2.4 & 73.3$\pm$6.1 & \underline{80.3$\pm$4.3} & 78.5$\pm$3.6 & \textbf{80.3$\pm$4.4} \\*
& AUC & 86.8$\pm$2.3 & 79.8$\pm$2.2 & 83.9$\pm$5.2 & 83.8$\pm$5.8 & 78.5$\pm$8.4 & 89.3$\pm$1.4 & 86.2$\pm$3.7 & \textbf{89.5$\pm$2.8} & 88.4$\pm$1.9 & \underline{89.3$\pm$3.1} \\
\multirow{3}{*}{HIF1A} & Pearson & 42.2$\pm$11.6 & 35.0$\pm$5.9 & 37.1$\pm$11.4 & 39.9$\pm$10.0 & 39.9$\pm$8.2 & 45.3$\pm$5.7 & 42.7$\pm$9.2 & \underline{48.4$\pm$10.0} & \textbf{49.4$\pm$7.4} & 47.8$\pm$10.0 \\*
& Spearman & 42.2$\pm$14.2 & 37.7$\pm$6.2 & 37.5$\pm$14.7 & 40.9$\pm$9.3 & 40.0$\pm$6.0 & 44.3$\pm$7.6 & 42.7$\pm$10.1 & \textbf{48.9$\pm$10.7} & \underline{48.6$\pm$7.9} & 48.3$\pm$12.0 \\*
& AUC & 71.9$\pm$9.9 & 69.0$\pm$3.6 & 68.9$\pm$9.1 & 71.9$\pm$6.0 & 69.8$\pm$7.1 & 70.1$\pm$3.8 & 71.7$\pm$6.6 & 74.3$\pm$3.3 & \underline{74.3$\pm$3.9} & \textbf{74.4$\pm$5.4} \\
\multirow{3}{*}{KEAP1} & Pearson & \underline{34.1$\pm$10.4} & 29.0$\pm$11.4 & 24.4$\pm$13.7 & 14.4$\pm$7.0 & 28.2$\pm$20.5 & 26.1$\pm$5.1 & 26.1$\pm$3.7 & 32.9$\pm$8.9 & 31.5$\pm$5.6 & \textbf{34.5$\pm$4.9} \\*
& Spearman & 33.1$\pm$7.1 & 25.6$\pm$12.3 & 24.2$\pm$8.4 & 15.2$\pm$5.4 & 25.8$\pm$22.5 & 26.2$\pm$5.1 & 25.4$\pm$7.6 & \underline{34.7$\pm$7.2} & 33.4$\pm$8.3 & \textbf{36.3$\pm$3.8} \\*
& AUC & 65.4$\pm$2.8 & 59.7$\pm$7.1 & 61.8$\pm$3.5 & 58.2$\pm$4.1 & 66.4$\pm$6.5 & 62.0$\pm$3.6 & 61.3$\pm$5.1 & \textbf{68.7$\pm$5.6} & 66.2$\pm$7.5 & \underline{67.9$\pm$4.2} \\
\multirow{3}{*}{KRAS} & Pearson & 20.4$\pm$8.0 & 14.5$\pm$9.3 & 15.4$\pm$3.8 & 19.6$\pm$10.3 & 9.2$\pm$19.9 & 27.8$\pm$6.7 & 22.1$\pm$10.7 & \underline{34.7$\pm$7.8} & \textbf{35.2$\pm$10.8} & 31.9$\pm$8.6 \\*
& Spearman & 20.3$\pm$8.0 & 15.3$\pm$8.6 & 17.8$\pm$3.1 & 20.4$\pm$7.7 & 11.7$\pm$24.0 & 33.3$\pm$7.0 & 23.6$\pm$8.3 & \textbf{38.3$\pm$8.6} & \underline{38.1$\pm$11.8} & 34.8$\pm$9.7 \\*
& AUC & 59.2$\pm$6.9 & 57.1$\pm$5.1 & 60.8$\pm$2.2 & 60.2$\pm$3.3 & 55.2$\pm$17.6 & 66.3$\pm$4.2 & 60.0$\pm$5.1 & \textbf{69.1$\pm$6.1} & \underline{68.7$\pm$6.2} & 66.4$\pm$5.0 \\
\multirow{3}{*}{LAG3} & Pearson & 70.4$\pm$2.1 & 64.9$\pm$5.9 & 62.2$\pm$7.1 & 60.1$\pm$6.0 & 46.6$\pm$11.1 & 69.5$\pm$6.5 & 70.5$\pm$3.8 & \underline{76.0$\pm$1.1} & 73.7$\pm$2.7 & \textbf{77.5$\pm$4.1} \\*
& Spearman & 67.5$\pm$2.2 & 62.5$\pm$7.8 & 59.6$\pm$4.6 & 56.1$\pm$7.4 & 45.2$\pm$13.4 & 69.5$\pm$7.0 & 68.9$\pm$5.0 & \underline{75.0$\pm$2.2} & 71.5$\pm$4.5 & \textbf{76.0$\pm$4.5} \\*
& AUC & 82.7$\pm$3.0 & 81.1$\pm$3.8 & 79.3$\pm$4.6 & 74.9$\pm$3.6 & 73.0$\pm$7.0 & 84.9$\pm$4.9 & 85.3$\pm$1.8 & \textbf{88.0$\pm$2.6} & 84.9$\pm$2.7 & \underline{87.6$\pm$1.9} \\
\multirow{3}{*}{MET} & Pearson & 33.9$\pm$7.6 & 30.0$\pm$6.9 & 31.7$\pm$11.8 & 22.1$\pm$13.6 & 36.7$\pm$14.9 & 41.5$\pm$6.7 & 36.5$\pm$6.6 & \textbf{46.2$\pm$7.5} & \underline{45.1$\pm$2.9} & 44.6$\pm$6.5 \\*
& Spearman & 37.8$\pm$12.6 & 29.4$\pm$8.4 & 31.9$\pm$14.0 & 24.7$\pm$17.0 & 38.4$\pm$16.3 & 41.9$\pm$9.9 & 34.9$\pm$6.4 & \textbf{46.5$\pm$4.7} & 45.7$\pm$2.6 & \underline{45.9$\pm$3.7} \\*
& AUC & 71.8$\pm$7.8 & 67.4$\pm$5.9 & 69.2$\pm$7.9 & 63.3$\pm$9.5 & 69.1$\pm$10.1 & 71.1$\pm$3.8 & 70.0$\pm$4.1 & \textbf{76.5$\pm$3.6} & 74.2$\pm$1.5 & \underline{75.0$\pm$3.1} \\
\multirow{3}{*}{MKI67} & Pearson & 54.2$\pm$3.3 & 51.4$\pm$6.0 & 35.2$\pm$6.7 & 38.2$\pm$8.1 & 41.3$\pm$11.8 & 50.1$\pm$9.7 & 48.2$\pm$4.2 & 57.0$\pm$4.5 & \underline{58.1$\pm$5.3} & \textbf{60.5$\pm$4.1} \\*
& Spearman & 44.9$\pm$9.1 & 42.8$\pm$8.3 & 29.2$\pm$4.9 & 33.3$\pm$11.5 & 39.7$\pm$8.7 & 47.3$\pm$10.7 & 43.4$\pm$9.9 & 52.5$\pm$6.5 & \underline{53.6$\pm$9.7} & \textbf{53.7$\pm$8.6} \\*
& AUC & 69.6$\pm$5.8 & 68.5$\pm$4.1 & 62.1$\pm$4.0 & 63.9$\pm$5.6 & 65.0$\pm$10.8 & 71.5$\pm$6.7 & 68.9$\pm$5.2 & 72.6$\pm$5.0 & \underline{72.7$\pm$6.8} & \textbf{73.1$\pm$6.5} \\
\multirow{3}{*}{PDCD1} & Pearson & 64.7$\pm$4.7 & 60.1$\pm$5.6 & 61.0$\pm$3.5 & 59.3$\pm$5.4 & 47.1$\pm$11.5 & 63.5$\pm$5.3 & 63.7$\pm$6.7 & \underline{70.4$\pm$5.0} & 69.9$\pm$2.8 & \textbf{72.9$\pm$6.2} \\*
& Spearman & 63.5$\pm$3.6 & 58.5$\pm$6.1 & 59.3$\pm$5.7 & 57.2$\pm$4.6 & 50.9$\pm$11.8 & 63.3$\pm$5.4 & 62.7$\pm$8.1 & \underline{69.1$\pm$4.5} & 69.0$\pm$2.1 & \textbf{72.0$\pm$6.1} \\*
& AUC & 82.2$\pm$2.9 & 78.7$\pm$3.9 & 79.9$\pm$4.3 & 79.6$\pm$1.9 & 75.2$\pm$7.3 & 80.5$\pm$4.4 & 80.2$\pm$5.1 & \underline{84.6$\pm$4.7} & 83.1$\pm$3.1 & \textbf{85.0$\pm$4.4} \\
\multirow{3}{*}{PTPRC} & Pearson & 65.7$\pm$3.0 & 61.6$\pm$7.7 & 60.6$\pm$2.9 & 59.5$\pm$4.7 & 46.0$\pm$10.9 & 68.3$\pm$4.5 & 69.4$\pm$7.8 & \underline{73.7$\pm$5.5} & 73.0$\pm$5.3 & \textbf{75.8$\pm$5.3} \\*
& Spearman & 66.9$\pm$5.3 & 61.4$\pm$9.3 & 63.2$\pm$2.5 & 60.8$\pm$5.7 & 44.3$\pm$10.4 & 69.0$\pm$5.9 & 70.2$\pm$8.6 & \underline{73.9$\pm$5.8} & 73.5$\pm$5.5 & \textbf{76.3$\pm$5.2} \\*
& AUC & 81.4$\pm$4.8 & 78.6$\pm$6.2 & 81.2$\pm$3.4 & 80.3$\pm$3.8 & 70.8$\pm$10.7 & 85.5$\pm$4.4 & 84.8$\pm$5.8 & \underline{85.7$\pm$4.9} & 85.2$\pm$5.3 & \textbf{87.5$\pm$5.0} \\
\multirow{3}{*}{STK11} & Pearson & 27.7$\pm$8.3 & 20.1$\pm$4.5 & 26.0$\pm$6.9 & 5.9$\pm$15.7 & \underline{31.2$\pm$9.9} & 23.4$\pm$12.1 & 20.5$\pm$9.1 & 28.1$\pm$11.0 & 30.1$\pm$4.9 & \textbf{33.5$\pm$7.2} \\*
& Spearman & 31.2$\pm$11.8 & 20.6$\pm$6.0 & 25.3$\pm$14.9 & 10.9$\pm$8.8 & \textbf{37.8$\pm$13.6} & 24.0$\pm$13.7 & 24.0$\pm$10.4 & 30.5$\pm$10.4 & 32.3$\pm$5.9 & \underline{36.3$\pm$7.9} \\*
& AUC & 66.1$\pm$8.9 & 60.5$\pm$3.9 & 62.1$\pm$9.4 & 56.0$\pm$1.6 & \underline{68.0$\pm$9.0} & 60.4$\pm$6.3 & 61.8$\pm$5.0 & 66.2$\pm$8.0 & 65.8$\pm$5.6 & \textbf{69.8$\pm$7.3} \\
\multirow{3}{*}{TP53} & Pearson & 6.7$\pm$7.1 & 7.8$\pm$10.0 & 5.1$\pm$14.2 & 11.0$\pm$10.7 & 17.9$\pm$34.1 & 13.0$\pm$5.4 & \textbf{23.6$\pm$10.2} & 16.8$\pm$13.1 & 19.0$\pm$8.6 & \underline{22.1$\pm$12.8} \\*
& Spearman & 7.7$\pm$1.4 & 7.9$\pm$6.5 & 3.9$\pm$15.5 & 12.3$\pm$7.4 & 14.5$\pm$31.3 & 16.3$\pm$5.2 & \textbf{23.9$\pm$5.1} & 20.1$\pm$9.2 & 19.5$\pm$5.9 & \underline{23.3$\pm$9.7} \\*
& AUC & 53.5$\pm$4.1 & 53.0$\pm$3.6 & 52.8$\pm$8.0 & 57.3$\pm$4.4 & 57.8$\pm$17.6 & 58.8$\pm$3.0 & \underline{61.7$\pm$4.0} & 61.2$\pm$3.9 & 60.1$\pm$2.0 & \textbf{62.7$\pm$3.8} \\
\multirow{3}{*}{VIM} & Pearson & 59.9$\pm$8.4 & 54.7$\pm$6.2 & 52.0$\pm$7.6 & 53.1$\pm$7.4 & 33.7$\pm$20.7 & \underline{67.9$\pm$5.9} & 64.0$\pm$2.2 & \textbf{68.0$\pm$6.1} & 67.4$\pm$5.9 & 67.5$\pm$4.9 \\*
& Spearman & 61.9$\pm$6.9 & 55.5$\pm$6.2 & 54.8$\pm$7.0 & 52.4$\pm$10.5 & 35.6$\pm$21.2 & \textbf{68.9$\pm$6.7} & 65.5$\pm$3.7 & 67.8$\pm$8.0 & 68.1$\pm$7.0 & \underline{68.2$\pm$6.6} \\*
& AUC & 82.8$\pm$2.3 & 78.4$\pm$6.2 & 78.8$\pm$2.5 & 75.5$\pm$6.8 & 70.4$\pm$12.6 & 85.7$\pm$4.0 & 84.9$\pm$2.3 & 85.5$\pm$4.1 & \textbf{85.7$\pm$3.4} & \underline{85.7$\pm$3.2} \\
\end{longtable}
\endgroup

\begingroup
\setlength{\tabcolsep}{1.15pt}
\renewcommand{\arraystretch}{1.02}
\setlength{\LTleft}{0pt}
\setlength{\LTright}{0pt}
\setlength{\LTcapwidth}{\textwidth}
\begin{longtable}{@{\extracolsep{\fill}}
>{\scriptsize}l
>{\scriptsize}l
*{10}{>{\scriptsize}c}
@{}}
\caption{Full gene-level prediction results on TCGA-BRCA. Each cell reports mean$\pm$std over five folds. The best and second-best results for each gene-metric pair are shown in bold and underlined, respectively.}\label{tab:gene_full_brca_all_models}\\
\toprule
Gene & Metric & ABMIL & DSMIL & TransMIL & RRT-MIL & TITAN & CHIEF & Feather & GECKO & ConcepPath & Ours \\
\midrule
\endfirsthead
\caption[]{Full gene-level prediction results on TCGA-BRCA (continued).}\\
\toprule
Gene & Metric & ABMIL & DSMIL & TransMIL & RRT-MIL & TITAN & CHIEF & Feather & GECKO & ConcepPath & Ours \\
\midrule
\endhead
\midrule
\multicolumn{12}{r}{\scriptsize Continued on next page}\\
\endfoot
\bottomrule
\endlastfoot
\multirow{3}{*}{ACTA2} & Pearson & 42.2$\pm$5.3 & 40.6$\pm$10.8 & 40.2$\pm$8.6 & 39.9$\pm$8.1 & 56.3$\pm$6.6 & 54.3$\pm$5.5 & 50.3$\pm$5.2 & 58.0$\pm$4.8 & \textbf{59.2$\pm$7.0} & \underline{58.1$\pm$6.4} \\*
& Spearman & 40.2$\pm$3.5 & 37.7$\pm$11.2 & 39.4$\pm$7.4 & 37.5$\pm$9.0 & 54.5$\pm$6.2 & 54.4$\pm$3.9 & 48.7$\pm$6.3 & \underline{58.4$\pm$4.0} & \textbf{58.7$\pm$7.2} & 57.6$\pm$7.0 \\*
& AUC & 70.1$\pm$2.3 & 68.8$\pm$6.2 & 70.0$\pm$3.9 & 68.9$\pm$5.1 & 77.2$\pm$3.3 & 77.4$\pm$3.6 & 73.8$\pm$3.8 & \underline{79.5$\pm$3.0} & \textbf{79.7$\pm$3.7} & 79.3$\pm$4.3 \\
\multirow{3}{*}{BRAF} & Pearson & 20.5$\pm$5.1 & 15.6$\pm$3.9 & 18.8$\pm$5.1 & 21.2$\pm$3.3 & 29.9$\pm$8.0 & 29.9$\pm$8.8 & 30.5$\pm$5.3 & 34.0$\pm$7.7 & \textbf{36.9$\pm$6.4} & \underline{35.8$\pm$6.5} \\*
& Spearman & 19.7$\pm$4.9 & 13.7$\pm$4.7 & 17.5$\pm$7.7 & 20.7$\pm$3.5 & 28.5$\pm$7.5 & 30.0$\pm$9.5 & 29.8$\pm$5.6 & 33.9$\pm$8.4 & \textbf{36.9$\pm$5.5} & \underline{35.1$\pm$6.5} \\*
& AUC & 59.8$\pm$2.6 & 56.0$\pm$2.6 & 58.5$\pm$4.7 & 61.7$\pm$1.6 & 64.7$\pm$4.6 & 66.1$\pm$4.2 & 64.8$\pm$3.8 & 66.4$\pm$3.7 & \textbf{68.9$\pm$4.1} & \underline{68.5$\pm$2.6} \\
\multirow{3}{*}{CD274} & Pearson & 45.1$\pm$7.5 & 41.9$\pm$5.8 & 41.9$\pm$7.6 & 41.4$\pm$8.3 & 58.3$\pm$4.8 & 54.2$\pm$3.6 & 54.5$\pm$6.5 & \underline{59.5$\pm$2.1} & \textbf{60.5$\pm$3.1} & 59.0$\pm$2.5 \\*
& Spearman & 44.7$\pm$4.3 & 41.3$\pm$5.6 & 40.2$\pm$8.5 & 41.1$\pm$5.7 & 56.1$\pm$5.5 & 51.5$\pm$5.1 & 52.7$\pm$5.8 & \underline{57.9$\pm$3.6} & \textbf{58.1$\pm$3.8} & 57.2$\pm$2.6 \\*
& AUC & 71.3$\pm$2.9 & 70.0$\pm$3.9 & 69.3$\pm$6.1 & 70.4$\pm$3.2 & 77.0$\pm$3.5 & 74.0$\pm$3.6 & 75.6$\pm$2.9 & \textbf{77.8$\pm$3.0} & \underline{77.6$\pm$2.8} & 76.9$\pm$1.7 \\
\multirow{3}{*}{CD4} & Pearson & 58.3$\pm$2.3 & 51.6$\pm$6.0 & 53.3$\pm$4.6 & 53.9$\pm$5.2 & 67.0$\pm$3.5 & 65.9$\pm$2.9 & 62.9$\pm$3.8 & \underline{70.4$\pm$3.6} & \textbf{71.0$\pm$2.9} & 70.2$\pm$3.8 \\*
& Spearman & 56.8$\pm$4.4 & 51.2$\pm$9.7 & 50.9$\pm$4.3 & 51.0$\pm$6.2 & 65.8$\pm$4.7 & 64.6$\pm$3.5 & 60.9$\pm$5.1 & 69.2$\pm$4.1 & \textbf{70.0$\pm$4.3} & \underline{69.7$\pm$4.8} \\*
& AUC & 77.7$\pm$4.8 & 75.3$\pm$7.1 & 75.5$\pm$3.1 & 73.9$\pm$3.1 & 82.4$\pm$4.8 & 81.5$\pm$3.3 & 79.6$\pm$4.3 & 83.7$\pm$3.9 & \underline{84.2$\pm$4.6} & \textbf{84.6$\pm$4.3} \\
\multirow{3}{*}{CD8A} & Pearson & 59.4$\pm$5.5 & 55.8$\pm$8.7 & 56.1$\pm$6.2 & 52.7$\pm$10.4 & 70.7$\pm$4.0 & 67.3$\pm$2.5 & 63.6$\pm$5.1 & 72.0$\pm$4.3 & \underline{72.6$\pm$3.7} & \textbf{73.9$\pm$3.0} \\*
& Spearman & 58.5$\pm$6.6 & 54.4$\pm$10.2 & 54.3$\pm$8.5 & 50.2$\pm$12.0 & 69.1$\pm$4.9 & 65.8$\pm$3.1 & 60.5$\pm$4.7 & 70.0$\pm$4.2 & \underline{70.4$\pm$4.2} & \textbf{71.8$\pm$3.1} \\*
& AUC & 78.9$\pm$3.1 & 77.0$\pm$4.6 & 76.7$\pm$4.7 & 74.4$\pm$5.8 & 83.9$\pm$2.1 & 81.5$\pm$1.8 & 79.9$\pm$1.0 & 84.3$\pm$3.0 & \underline{84.9$\pm$2.2} & \textbf{85.5$\pm$1.8} \\
\multirow{3}{*}{CDH1} & Pearson & 57.9$\pm$6.4 & 54.4$\pm$7.1 & 52.0$\pm$5.8 & 51.5$\pm$5.8 & 59.4$\pm$8.4 & 61.2$\pm$4.6 & 60.5$\pm$6.8 & 63.8$\pm$5.3 & \textbf{65.7$\pm$4.7} & \underline{64.5$\pm$5.5} \\*
& Spearman & 47.7$\pm$3.9 & 44.3$\pm$6.3 & 45.1$\pm$5.9 & 42.1$\pm$6.6 & 51.1$\pm$9.0 & 52.6$\pm$5.9 & 50.9$\pm$6.5 & 55.0$\pm$4.0 & \textbf{56.5$\pm$7.5} & \underline{56.4$\pm$6.9} \\*
& AUC & 70.7$\pm$2.5 & 69.0$\pm$3.9 & 69.8$\pm$4.1 & 69.0$\pm$4.6 & 73.7$\pm$5.7 & 74.6$\pm$4.2 & 73.1$\pm$4.6 & 75.5$\pm$3.7 & \textbf{76.5$\pm$5.4} & \underline{76.3$\pm$5.2} \\
\multirow{3}{*}{CTLA4} & Pearson & 65.2$\pm$3.0 & 60.8$\pm$1.0 & 61.9$\pm$4.2 & 61.8$\pm$2.7 & \underline{73.6$\pm$1.5} & 67.7$\pm$3.1 & 65.4$\pm$2.6 & 71.6$\pm$2.7 & 72.6$\pm$2.8 & \textbf{74.2$\pm$1.5} \\*
& Spearman & 64.0$\pm$6.9 & 59.4$\pm$4.3 & 59.1$\pm$4.8 & 60.5$\pm$5.9 & \underline{72.3$\pm$3.3} & 65.8$\pm$4.2 & 62.1$\pm$3.3 & 69.3$\pm$3.9 & 70.7$\pm$3.4 & \textbf{73.0$\pm$2.7} \\*
& AUC & 81.4$\pm$4.8 & 78.8$\pm$3.5 & 78.1$\pm$2.9 & 79.0$\pm$4.5 & \underline{86.1$\pm$3.1} & 83.1$\pm$2.6 & 80.8$\pm$3.0 & 84.0$\pm$2.9 & 84.8$\pm$2.7 & \textbf{86.4$\pm$2.3} \\
\multirow{3}{*}{CXCL10} & Pearson & 61.1$\pm$2.4 & 57.4$\pm$4.6 & 56.8$\pm$2.2 & 56.4$\pm$4.9 & \underline{70.5$\pm$2.1} & 65.9$\pm$4.9 & 64.4$\pm$2.3 & 69.9$\pm$3.5 & 70.4$\pm$3.6 & \textbf{70.9$\pm$2.3} \\*
& Spearman & 58.7$\pm$3.5 & 56.2$\pm$7.5 & 55.0$\pm$1.6 & 54.6$\pm$6.9 & \underline{70.4$\pm$2.8} & 64.3$\pm$6.0 & 62.1$\pm$4.1 & 69.6$\pm$4.6 & 69.6$\pm$4.5 & \textbf{70.8$\pm$2.9} \\*
& AUC & 77.7$\pm$2.7 & 77.0$\pm$4.2 & 75.4$\pm$1.1 & 75.6$\pm$5.6 & \underline{84.5$\pm$1.4} & 80.8$\pm$2.2 & 79.7$\pm$2.2 & 84.3$\pm$2.1 & 84.4$\pm$2.6 & \textbf{84.7$\pm$2.0} \\
\multirow{3}{*}{CXCL9} & Pearson & 62.0$\pm$5.1 & 58.0$\pm$7.0 & 61.3$\pm$6.5 & 61.4$\pm$9.0 & 70.2$\pm$3.2 & 66.2$\pm$3.5 & 62.8$\pm$2.3 & 70.6$\pm$3.8 & \underline{71.5$\pm$2.7} & \textbf{72.0$\pm$2.7} \\*
& Spearman & 61.0$\pm$6.2 & 57.1$\pm$9.9 & 59.8$\pm$7.6 & 60.5$\pm$10.1 & 69.2$\pm$3.8 & 64.0$\pm$3.4 & 60.3$\pm$2.2 & 69.0$\pm$4.1 & \underline{70.0$\pm$3.0} & \textbf{70.8$\pm$2.4} \\*
& AUC & 80.4$\pm$3.1 & 77.3$\pm$5.5 & 79.0$\pm$2.8 & 79.5$\pm$6.0 & 84.4$\pm$2.5 & 81.6$\pm$1.5 & 79.8$\pm$2.1 & 84.5$\pm$2.0 & \underline{84.8$\pm$1.5} & \textbf{85.4$\pm$1.5} \\
\multirow{3}{*}{EGFR} & Pearson & 52.4$\pm$6.5 & 47.9$\pm$9.9 & 43.3$\pm$9.6 & 49.0$\pm$6.3 & 57.8$\pm$4.4 & 52.4$\pm$7.4 & 52.3$\pm$2.9 & \underline{60.4$\pm$5.4} & \textbf{61.5$\pm$4.3} & 59.5$\pm$6.5 \\*
& Spearman & 54.0$\pm$6.3 & 50.4$\pm$9.5 & 44.5$\pm$8.7 & 50.1$\pm$7.0 & 57.9$\pm$4.3 & 50.5$\pm$6.9 & 51.6$\pm$2.7 & \underline{61.2$\pm$3.6} & \textbf{61.7$\pm$4.2} & 60.3$\pm$6.0 \\*
& AUC & 77.6$\pm$4.3 & 76.2$\pm$4.6 & 71.7$\pm$4.6 & 75.2$\pm$5.2 & 79.5$\pm$3.5 & 75.1$\pm$3.9 & 76.8$\pm$2.6 & \underline{81.5$\pm$2.8} & \textbf{81.5$\pm$3.1} & 80.8$\pm$3.7 \\
\multirow{3}{*}{ERBB2} & Pearson & 26.5$\pm$7.9 & 25.9$\pm$8.4 & 20.1$\pm$7.6 & 22.7$\pm$8.0 & 35.6$\pm$7.4 & 36.2$\pm$7.1 & 31.4$\pm$9.5 & 38.2$\pm$6.0 & \textbf{42.1$\pm$4.8} & \underline{41.2$\pm$6.5} \\*
& Spearman & 18.2$\pm$8.1 & 19.8$\pm$10.8 & 13.9$\pm$8.8 & 17.4$\pm$12.4 & 32.1$\pm$6.4 & 30.4$\pm$6.8 & 29.5$\pm$5.6 & 33.3$\pm$7.0 & \textbf{37.2$\pm$4.0} & \underline{35.6$\pm$3.5} \\*
& AUC & 56.0$\pm$4.6 & 57.1$\pm$7.1 & 54.3$\pm$4.8 & 55.7$\pm$7.0 & 62.3$\pm$3.5 & 62.4$\pm$5.0 & 62.2$\pm$1.9 & 64.6$\pm$2.7 & \textbf{66.4$\pm$1.0} & \underline{65.0$\pm$0.7} \\
\multirow{3}{*}{FOXP3} & Pearson & 59.9$\pm$2.9 & 54.3$\pm$5.0 & 57.3$\pm$5.7 & 57.1$\pm$6.6 & \underline{69.8$\pm$2.9} & 63.3$\pm$4.1 & 61.3$\pm$3.9 & 68.1$\pm$1.9 & 69.0$\pm$5.3 & \textbf{70.2$\pm$2.4} \\*
& Spearman & 59.7$\pm$3.6 & 54.4$\pm$7.6 & 56.8$\pm$7.3 & 55.3$\pm$8.0 & \underline{69.3$\pm$2.2} & 62.7$\pm$4.6 & 59.5$\pm$4.2 & 67.9$\pm$2.2 & 69.1$\pm$4.7 & \textbf{70.3$\pm$2.8} \\*
& AUC & 80.9$\pm$2.4 & 77.1$\pm$5.6 & 79.7$\pm$4.3 & 77.8$\pm$5.4 & 85.1$\pm$2.1 & 82.4$\pm$3.1 & 80.0$\pm$3.6 & 85.4$\pm$1.6 & \underline{85.6$\pm$3.5} & \textbf{86.6$\pm$2.4} \\
\multirow{3}{*}{HAVCR2} & Pearson & 53.4$\pm$1.5 & 48.4$\pm$4.3 & 45.1$\pm$3.8 & 49.6$\pm$4.2 & 61.6$\pm$5.1 & 61.7$\pm$6.8 & 58.3$\pm$4.6 & \textbf{66.4$\pm$4.0} & 65.3$\pm$5.4 & \underline{65.7$\pm$4.8} \\*
& Spearman & 51.1$\pm$2.3 & 45.9$\pm$5.0 & 41.8$\pm$4.1 & 47.5$\pm$2.4 & 61.2$\pm$5.6 & 60.4$\pm$7.1 & 57.6$\pm$6.3 & \underline{66.0$\pm$4.5} & 65.0$\pm$5.7 & \textbf{66.1$\pm$4.9} \\*
& AUC & 74.2$\pm$2.3 & 72.8$\pm$3.0 & 69.8$\pm$3.3 & 73.3$\pm$2.5 & 80.1$\pm$3.3 & 81.0$\pm$4.3 & 78.8$\pm$4.6 & \underline{82.4$\pm$2.6} & 81.9$\pm$3.6 & \textbf{82.6$\pm$3.4} \\
\multirow{3}{*}{HIF1A} & Pearson & 38.9$\pm$6.9 & 39.2$\pm$7.9 & 35.5$\pm$12.3 & 38.0$\pm$7.4 & 44.5$\pm$4.8 & 45.4$\pm$4.9 & 41.7$\pm$7.2 & 45.3$\pm$6.7 & \underline{46.4$\pm$5.1} & \textbf{47.6$\pm$7.1} \\*
& Spearman & 38.0$\pm$5.4 & 38.5$\pm$6.8 & 35.3$\pm$11.7 & 37.3$\pm$7.5 & 43.9$\pm$5.8 & 44.0$\pm$5.3 & 40.7$\pm$8.1 & 44.4$\pm$6.3 & \underline{44.5$\pm$5.7} & \textbf{44.9$\pm$7.1} \\*
& AUC & 67.7$\pm$2.5 & 68.0$\pm$2.5 & 66.9$\pm$5.9 & 67.9$\pm$4.1 & 71.9$\pm$4.7 & 71.8$\pm$2.7 & 70.4$\pm$4.3 & \underline{72.3$\pm$3.7} & 72.2$\pm$3.9 & \textbf{72.5$\pm$4.2} \\
\multirow{3}{*}{KEAP1} & Pearson & 20.3$\pm$3.8 & 22.7$\pm$3.0 & 18.8$\pm$8.3 & 19.9$\pm$2.9 & 22.6$\pm$8.4 & 27.3$\pm$5.3 & 21.2$\pm$5.0 & 31.5$\pm$7.2 & \textbf{37.7$\pm$6.6} & \underline{33.2$\pm$8.8} \\*
& Spearman & 19.6$\pm$4.0 & 21.8$\pm$3.1 & 17.6$\pm$7.1 & 20.0$\pm$3.7 & 24.3$\pm$7.8 & 26.4$\pm$4.8 & 22.9$\pm$6.1 & \underline{30.1$\pm$6.2} & \textbf{35.1$\pm$6.4} & 29.8$\pm$8.0 \\*
& AUC & 61.4$\pm$3.2 & 62.1$\pm$2.2 & 60.0$\pm$4.8 & 62.0$\pm$2.4 & 62.4$\pm$3.8 & 64.0$\pm$2.4 & 62.0$\pm$3.1 & \underline{65.2$\pm$3.7} & \textbf{66.4$\pm$5.3} & 64.4$\pm$3.9 \\
\multirow{3}{*}{KRAS} & Pearson & 16.5$\pm$5.4 & 14.5$\pm$4.8 & 19.4$\pm$6.4 & 18.0$\pm$4.1 & 26.0$\pm$4.3 & 27.6$\pm$2.7 & 18.9$\pm$5.5 & 27.8$\pm$2.8 & \textbf{30.3$\pm$2.8} & \underline{30.1$\pm$1.2} \\*
& Spearman & 17.5$\pm$6.1 & 11.9$\pm$4.3 & 17.2$\pm$5.8 & 16.4$\pm$4.9 & 26.7$\pm$4.2 & 27.9$\pm$4.8 & 17.2$\pm$2.8 & 28.9$\pm$4.5 & \textbf{30.5$\pm$4.4} & \underline{29.1$\pm$2.8} \\*
& AUC & 58.7$\pm$3.8 & 55.9$\pm$3.0 & 57.4$\pm$3.1 & 58.0$\pm$3.8 & 62.5$\pm$2.4 & 63.2$\pm$3.2 & 56.5$\pm$1.6 & \underline{64.1$\pm$3.3} & \textbf{64.2$\pm$3.5} & 63.5$\pm$2.1 \\
\multirow{3}{*}{LAG3} & Pearson & 60.2$\pm$2.6 & 57.9$\pm$3.6 & 54.5$\pm$6.1 & 56.9$\pm$2.3 & 66.5$\pm$2.3 & 64.3$\pm$2.5 & 61.4$\pm$1.7 & \underline{68.0$\pm$0.9} & 67.9$\pm$2.2 & \textbf{70.1$\pm$0.6} \\*
& Spearman & 55.4$\pm$5.0 & 54.0$\pm$6.7 & 47.8$\pm$6.1 & 51.9$\pm$4.1 & 62.8$\pm$4.2 & 60.2$\pm$2.3 & 56.1$\pm$3.2 & \underline{63.9$\pm$3.5} & 63.0$\pm$3.1 & \textbf{66.1$\pm$3.1} \\*
& AUC & 77.0$\pm$4.0 & 76.1$\pm$4.9 & 73.2$\pm$3.1 & 75.9$\pm$1.9 & 80.6$\pm$4.4 & 78.4$\pm$1.1 & 76.8$\pm$2.6 & \underline{81.9$\pm$2.7} & 81.1$\pm$3.1 & \textbf{82.1$\pm$3.1} \\
\multirow{3}{*}{MET} & Pearson & 47.6$\pm$5.0 & 41.4$\pm$8.2 & 42.9$\pm$7.2 & 45.9$\pm$4.2 & 54.3$\pm$4.6 & 50.9$\pm$5.5 & 51.5$\pm$5.5 & 57.1$\pm$3.1 & \textbf{59.4$\pm$3.2} & \underline{59.0$\pm$3.8} \\*
& Spearman & 48.2$\pm$4.9 & 42.1$\pm$7.8 & 43.0$\pm$6.4 & 46.9$\pm$4.6 & 55.1$\pm$2.8 & 50.6$\pm$3.9 & 51.6$\pm$4.5 & 57.8$\pm$2.3 & \textbf{60.1$\pm$3.0} & \underline{59.5$\pm$3.2} \\*
& AUC & 74.1$\pm$4.2 & 70.7$\pm$4.5 & 71.1$\pm$3.3 & 72.9$\pm$4.8 & 77.4$\pm$2.0 & 74.7$\pm$2.5 & 75.9$\pm$2.3 & 79.5$\pm$1.9 & \underline{80.5$\pm$2.6} & \textbf{80.7$\pm$2.7} \\
\multirow{3}{*}{MKI67} & Pearson & 58.1$\pm$3.1 & 51.2$\pm$9.0 & 55.1$\pm$5.7 & 52.9$\pm$7.4 & 60.6$\pm$4.0 & 59.1$\pm$4.6 & 57.7$\pm$0.8 & 64.9$\pm$2.0 & \underline{65.3$\pm$2.5} & \textbf{65.4$\pm$1.4} \\*
& Spearman & 60.3$\pm$2.0 & 52.7$\pm$7.1 & 55.8$\pm$6.2 & 54.0$\pm$8.3 & 62.2$\pm$3.3 & 60.1$\pm$4.2 & 58.4$\pm$1.8 & 65.7$\pm$1.6 & \underline{66.0$\pm$2.9} & \textbf{66.4$\pm$1.6} \\*
& AUC & 80.3$\pm$1.3 & 76.8$\pm$2.2 & 77.5$\pm$4.1 & 76.4$\pm$6.1 & 81.7$\pm$2.7 & 80.8$\pm$3.0 & 78.6$\pm$3.7 & 82.7$\pm$2.0 & \textbf{83.0$\pm$2.9} & \underline{83.0$\pm$2.6} \\
\multirow{3}{*}{PDCD1} & Pearson & 61.4$\pm$3.4 & 54.4$\pm$11.6 & 56.4$\pm$5.7 & 58.7$\pm$4.6 & 71.6$\pm$2.3 & 66.9$\pm$3.7 & 63.9$\pm$3.5 & 71.0$\pm$3.2 & \underline{72.9$\pm$3.8} & \textbf{73.6$\pm$2.3} \\*
& Spearman & 58.8$\pm$5.7 & 52.0$\pm$13.6 & 52.3$\pm$8.1 & 55.1$\pm$7.0 & 69.7$\pm$2.3 & 64.6$\pm$4.5 & 60.4$\pm$3.9 & 68.6$\pm$3.7 & \underline{71.1$\pm$4.8} & \textbf{71.8$\pm$2.9} \\*
& AUC & 78.8$\pm$3.4 & 75.2$\pm$6.6 & 75.2$\pm$4.5 & 76.6$\pm$4.7 & 85.0$\pm$1.2 & 81.2$\pm$2.7 & 79.9$\pm$1.8 & 84.5$\pm$0.8 & \textbf{86.0$\pm$1.4} & \underline{85.2$\pm$1.2} \\
\multirow{3}{*}{PTPRC} & Pearson & 59.7$\pm$4.3 & 51.5$\pm$11.1 & 56.5$\pm$4.9 & 54.0$\pm$6.5 & 69.7$\pm$3.5 & 67.2$\pm$2.8 & 64.2$\pm$4.5 & \underline{71.6$\pm$3.4} & \textbf{72.2$\pm$2.1} & 71.5$\pm$3.5 \\*
& Spearman & 58.6$\pm$6.1 & 51.2$\pm$12.8 & 54.2$\pm$6.1 & 53.0$\pm$6.8 & 68.6$\pm$4.7 & 65.6$\pm$3.6 & 62.7$\pm$4.8 & \underline{71.3$\pm$4.3} & \textbf{71.7$\pm$2.8} & 71.3$\pm$4.3 \\*
& AUC & 78.4$\pm$3.6 & 75.5$\pm$6.3 & 76.2$\pm$3.7 & 75.6$\pm$3.8 & 82.3$\pm$3.6 & 81.1$\pm$2.1 & 80.7$\pm$3.0 & \textbf{84.9$\pm$3.1} & \underline{84.7$\pm$2.5} & 84.0$\pm$2.5 \\
\multirow{3}{*}{STK11} & Pearson & 31.4$\pm$7.4 & 26.6$\pm$6.9 & 27.4$\pm$2.9 & 24.0$\pm$2.6 & 35.6$\pm$6.1 & 37.3$\pm$3.1 & 33.9$\pm$7.7 & 41.7$\pm$2.9 & \textbf{46.6$\pm$5.4} & \underline{43.6$\pm$3.5} \\*
& Spearman & 26.5$\pm$7.7 & 24.6$\pm$8.0 & 25.8$\pm$6.6 & 23.6$\pm$3.7 & 32.3$\pm$5.3 & 35.0$\pm$3.9 & 30.7$\pm$5.2 & \underline{40.0$\pm$4.5} & \textbf{43.2$\pm$3.9} & 39.9$\pm$3.4 \\*
& AUC & 62.1$\pm$4.3 & 61.7$\pm$4.8 & 61.4$\pm$3.4 & 61.9$\pm$3.7 & 64.7$\pm$3.8 & 67.4$\pm$1.1 & 64.6$\pm$3.3 & 69.8$\pm$3.4 & \textbf{71.6$\pm$2.2} & \underline{70.3$\pm$2.4} \\
\multirow{3}{*}{TP53} & Pearson & 8.1$\pm$5.2 & 5.3$\pm$4.8 & 6.6$\pm$6.8 & 3.9$\pm$7.0 & 11.4$\pm$8.5 & 9.9$\pm$5.5 & 9.7$\pm$4.8 & 10.8$\pm$7.9 & \underline{11.5$\pm$7.3} & \textbf{11.8$\pm$5.7} \\*
& Spearman & 3.1$\pm$6.9 & 5.2$\pm$5.3 & 6.5$\pm$5.6 & 6.1$\pm$4.8 & 9.0$\pm$7.1 & 7.3$\pm$3.8 & 7.5$\pm$6.2 & 8.9$\pm$4.3 & \textbf{11.2$\pm$5.3} & \underline{9.8$\pm$5.4} \\*
& AUC & 50.3$\pm$4.0 & 52.8$\pm$2.5 & 53.7$\pm$3.5 & 54.6$\pm$3.3 & 54.5$\pm$3.2 & 54.2$\pm$1.3 & \underline{55.3$\pm$2.0} & 54.4$\pm$3.9 & \textbf{56.3$\pm$2.8} & 54.7$\pm$1.7 \\
\multirow{3}{*}{VIM} & Pearson & 49.2$\pm$4.9 & 44.0$\pm$11.4 & 42.1$\pm$7.2 & 43.9$\pm$7.4 & 56.0$\pm$6.5 & 55.3$\pm$5.0 & 49.7$\pm$1.5 & 57.9$\pm$3.9 & \underline{58.9$\pm$4.6} & \textbf{59.6$\pm$4.2} \\*
& Spearman & 48.0$\pm$3.1 & 43.3$\pm$9.5 & 40.3$\pm$6.6 & 42.3$\pm$9.0 & 54.4$\pm$6.6 & 54.4$\pm$5.3 & 47.0$\pm$2.4 & 56.9$\pm$4.6 & \underline{58.1$\pm$5.2} & \textbf{58.9$\pm$4.6} \\*
& AUC & 73.7$\pm$2.5 & 71.7$\pm$4.9 & 70.2$\pm$3.3 & 70.9$\pm$4.3 & 77.0$\pm$1.8 & 77.3$\pm$2.9 & 72.5$\pm$1.5 & 78.3$\pm$2.5 & \underline{79.1$\pm$2.8} & \textbf{79.9$\pm$1.7} \\
\end{longtable}
\endgroup

\section{Additional Visualization Results}
\label{app:visualization_results}

\subsection{Five-fold Kaplan--Meier Survival Curves}
\label{app:km_curves}

\begin{figure*}[!t]
\centering
\includegraphics[width=\textwidth]{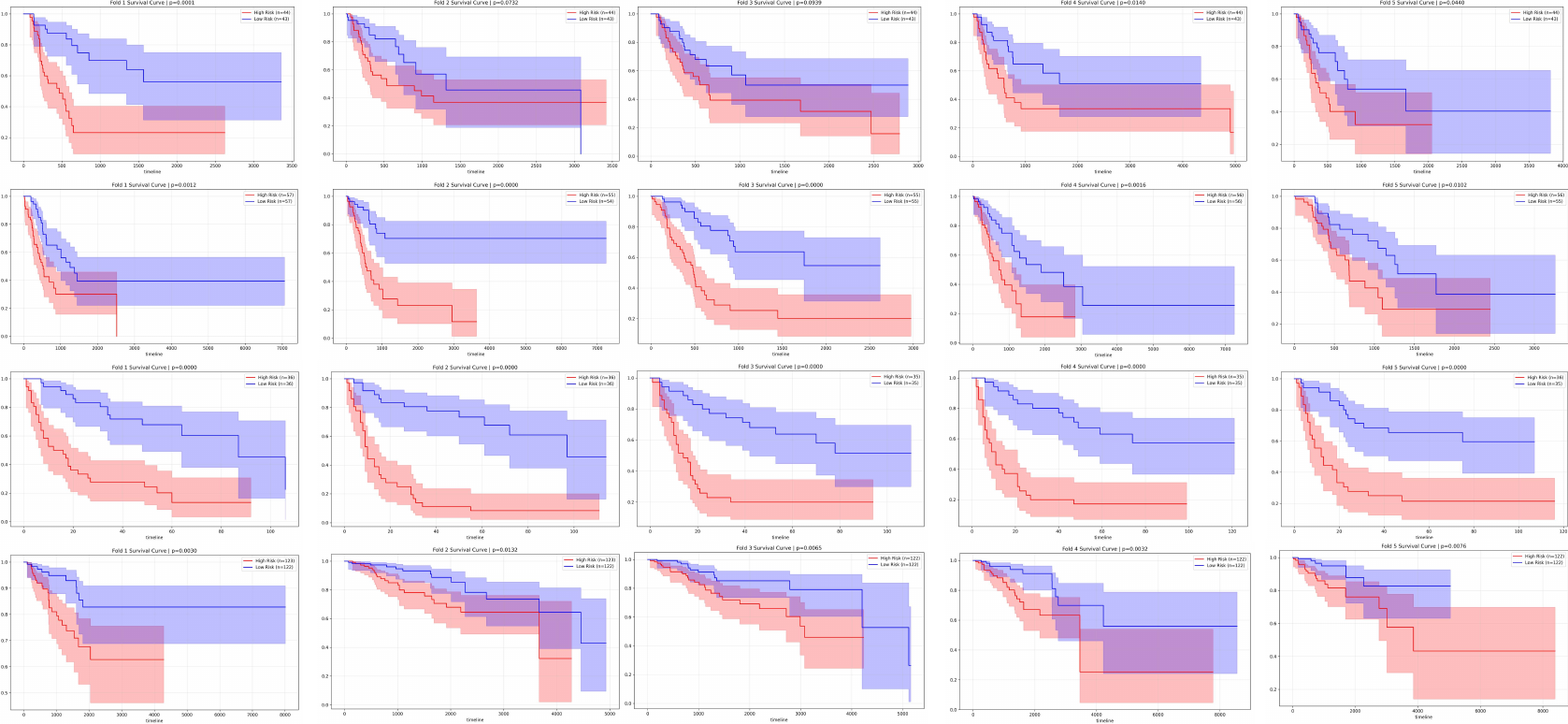}
\caption{
Five-fold Kaplan--Meier survival curves for survival prediction.
Rows from top to bottom correspond to BLCA, LUAD, GBC, and BRCA, respectively.
For each dataset, the five columns show the Kaplan--Meier curves from the five cross-validation folds.
Patients are stratified into high-risk and low-risk groups according to the predicted slide-level risk scores.
}
\label{fig:appendix_km_curves}
\vspace{-2mm}
\end{figure*}

To further evaluate the survival stratification ability of TMEvolve, we provide the Kaplan--Meier survival curves for all five cross-validation folds. 
For each fold, patients are divided into high-risk and low-risk groups according to the predicted risk scores. 
As shown in Figure~\ref{fig:appendix_km_curves}, TMEvolve generally separates the two risk groups across different cohorts, indicating that the learned slide-level risk representations are consistently associated with patient survival outcomes.

\subsection{Additional Interpretability Visualization}
\label{app:additional_interpretability}

\begin{figure*}[!t]
\centering
\includegraphics[width=\textwidth]{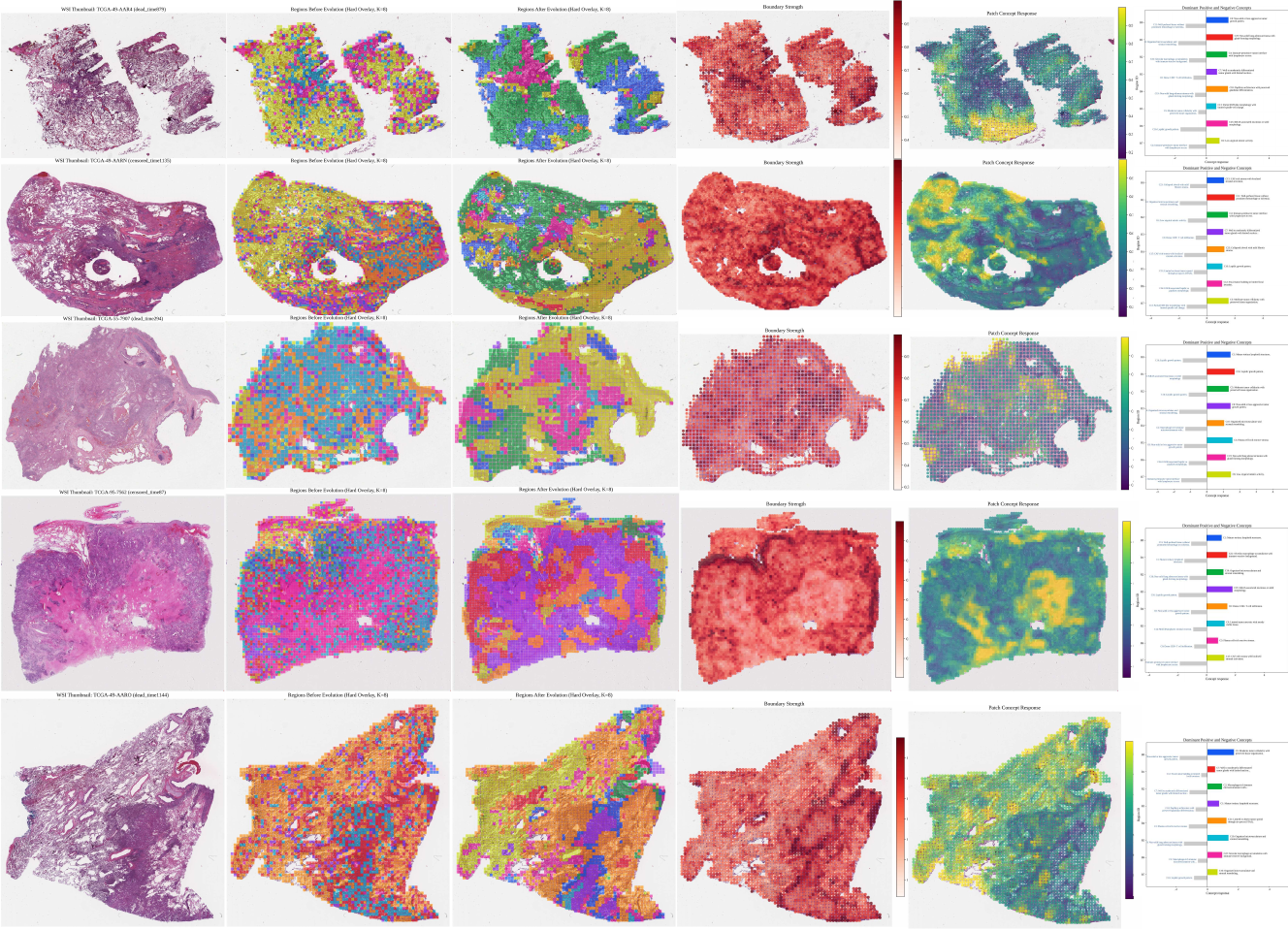}
\caption{
Additional interpretability visualization of TMEvolve on representative TCGA-LUAD WSI cases.
From left to right, we show the original WSI thumbnail, the region partition before evolution, the region partition after evolution, the boundary strength map, the patch-level concept response map, and the dominant concept response summary.
Compared with the initial partitions, the post-evolution regions become more spatially coherent, while the boundary and concept response maps highlight heterogeneous tissue interfaces and concept-related microenvironmental patterns in LUAD.
}
\label{fig:appendix_interpretability_vis}
\vspace{-2mm}
\end{figure*}

To provide more qualitative evidence for the interpretability of TMEvolve, we present additional visualization results on representative WSI cases in Figure~\ref{fig:appendix_interpretability_vis}. 
For each case, we visualize the original WSI thumbnail, the region partition before evolution, the region partition after evolution, the boundary strength map, the patch-level concept response map, and the dominant concept response summary. 
These visualizations show how TMEvolve organizes patch-level features into coherent tissue regions and further refines the region structure through concept-guided microenvironment evolution.

\section{Limitations}
\label{app:limitations}

TMEvolve is designed as a representation-learning framework for modeling region-level microenvironment organization, rather than a mechanistic simulation of biological tumor progression. 
The pseudo-time evolution process should therefore be interpreted as iterative feature refinement over heterogeneous tissue regions, not as true temporal dynamics of the tumor microenvironment. 
In addition, although the pathological concept bank is reviewed by experts for histopathological relevance and visual observability in H\&E WSIs, its coverage may still be limited for rare or dataset-specific patterns.

Another limitation is that region partitioning is learned from slide-level supervision without pixel-level tissue annotations. 
Thus, the learned regions are model-derived microenvironment organizations rather than exact pathological segmentation masks. 
While our visualizations suggest improved spatial coherence and conceptual interpretability, further validation using expert annotations, spatial molecular profiles, or multi-scale region graphs would strengthen the biological interpretation and extend TMEvolve to capture longer-range tissue interactions.

\section{Broader Impact}
\label{app:broader_impact}

TMEvolve may have positive societal impacts by improving weakly supervised computational pathology models for prognosis, molecular prediction, and subtype classification, and by providing region- and boundary-level interpretability that may support pathology research and clinical decision support. 
However, the method also carries potential negative impacts if deployed without sufficient validation. 
Prediction errors, dataset bias, domain shifts across hospitals or scanners, and over-interpretation of model-derived regions may lead to misleading conclusions in clinical settings. 
Therefore, TMEvolve should be used as a research and decision-support tool rather than a standalone diagnostic system, and any clinical deployment would require rigorous external validation, expert oversight, privacy protection, and compliance with institutional and ethical regulations.

\clearpage

\end{document}